\documentclass{egpubl}
\usepackage{VMV2023}
 
\WsSubmission      % uncomment for submission to EG Workshop
\usepackage{amssymb}
\usepackage{subcaption}
\usepackage[T1]{fontenc}
\usepackage{dfadobe}
\usepackage{amsmath}
\DeclareMathOperator{\cov}{cov}
\DeclareMathOperator{\var}{var}
\DeclareMathOperator{\enc}{Enc}
\DeclareMathOperator{\grid}{Grid}
\DeclareMathOperator*{\argmin}{arg\,min}
\usepackage{cite}  % comment out for biblatex with backend=biber
\BibtexOrBiblatex
\electronicVersion
\PrintedOrElectronic
\ifpdf \usepackage[pdftex]{graphicx} \pdfcompresslevel=9
\else \usepackage[dvips]{graphicx} \fi

\usepackage{egweblnk} 
\graphicspath{{figs/}{figures/}{pictures/}{images/}{./}}
\title[Neural Fields for Interactive Visualization of Statistical Dependencies in 3D Simulation Ensembles]%
      {Neural Fields for Interactive Visualization of Statistical Dependencies in 3D Simulation Ensembles}

\author[F. Farokhmanesh et al.]
{\parbox{\textwidth}{\centering F. Farokhmanesh$^{1}$\orcid{0000-0002-3595-1629}, K.              Höhlein$^{1}$\orcid{0000-0002-4483-8388}, C. Neuhauser$^{1}$\orcid{0000-0002-0290-1991}, T. Necker$^{2}$\orcid{0000-0002-7484-3372}, M. Weissmann$^{2}$\orcid{0000-0003-4073-1791}, T. Miyoshi$^{3}$\orcid{0000-0003-3160-2525}
        and R. Westermann$^{1}$\orcid{0000-0002-3394-0731} 
        }
        \\
{\parbox{\textwidth}{\centering $^1$Technical University of Munich, Department of Computer Science, School of Computation, Information and Technology, Germany 
        \\$^2$  University of Vienna, Department of Meteorology and Geophysics, Austria
        \\$^3$  RIKEN Center for Computational Science, Kobe, Japan
        }
        }
        }
\begin{document}
% uncomment for using teaser
\teaser{
  \centering
  \begin{subfigure}[b]{0.32\columnwidth}
  	\centering
  	\includegraphics[width=\textwidth]{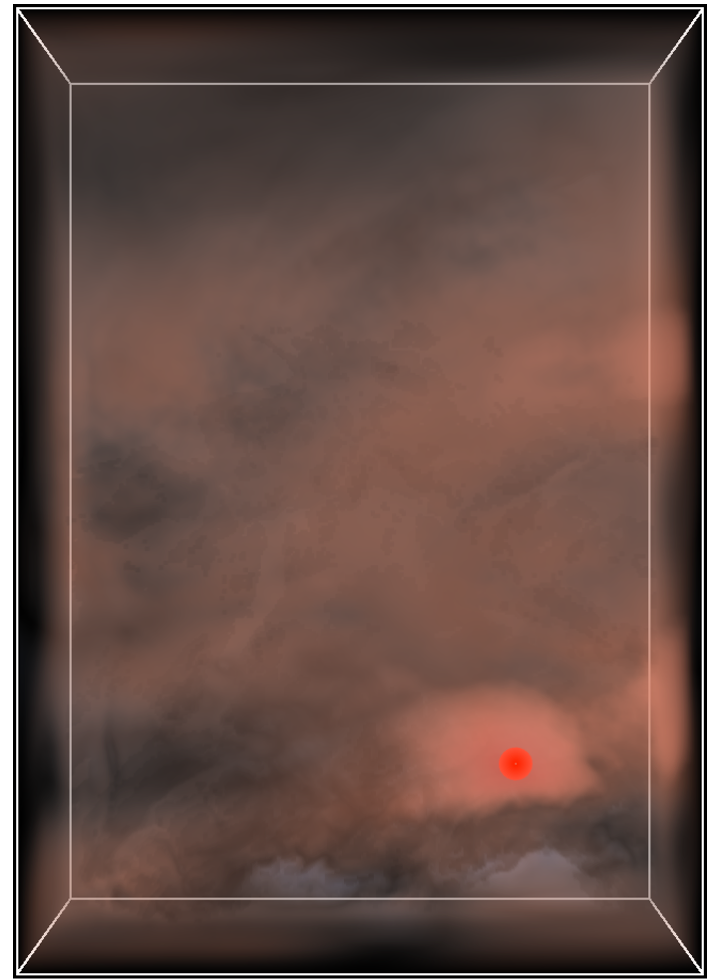}
  	\caption{Ground truth}
  	\label{fig:teaser:pearson:gt}
  \end{subfigure}%
  \begin{subfigure}[b]{0.32\columnwidth}
  	\centering
  	\includegraphics[width=\textwidth]{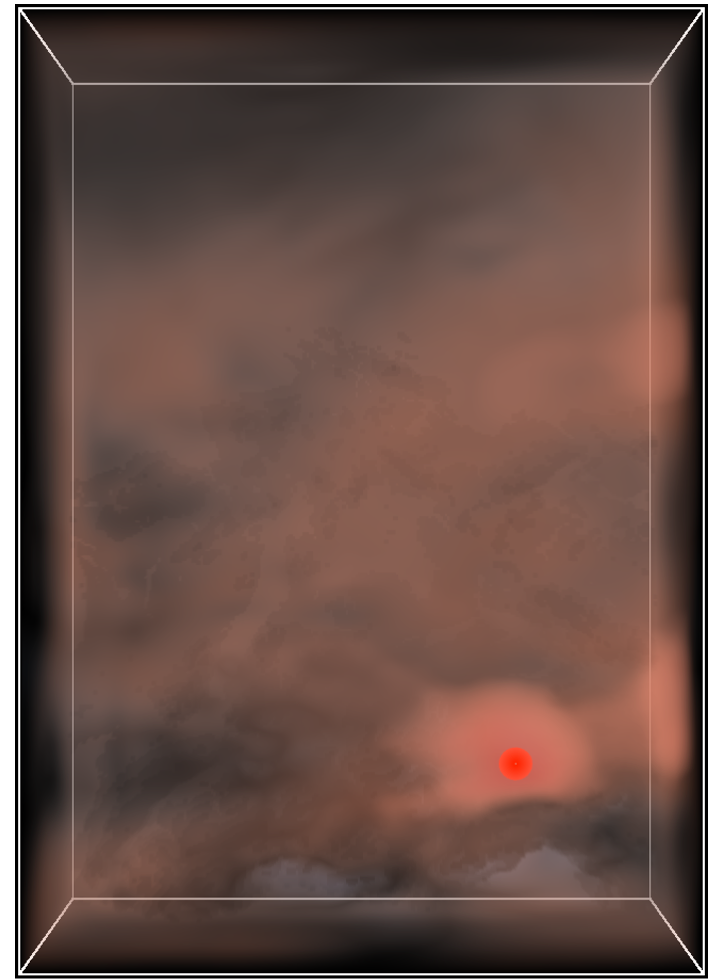}
  	\caption{NDF reconstruction}
  	\label{fig:teaser:pearson:ndf}
  \end{subfigure}%
  \begin{subfigure}[b]{0.07\columnwidth}
  	\centering
  	\includegraphics[width=\textwidth]{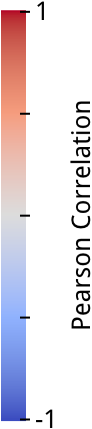}
  	\parbox{\textwidth}{~}
%   	\caption{}
  	\label{fig:teaser:pearson:tf}
  \end{subfigure}%
  \hspace{0.8cm}
    \begin{subfigure}[b]{0.32\columnwidth}
  	\centering
  	\includegraphics[width=\textwidth]{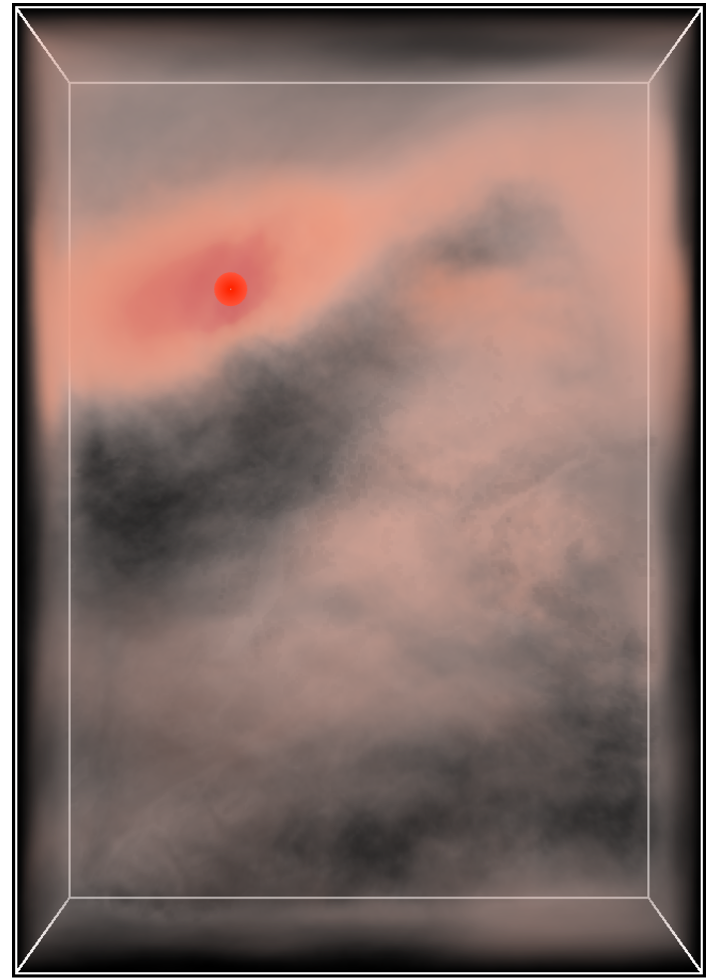}
  	\caption{Ground truth}
  	\label{fig:teaser:mi:gt}
  \end{subfigure}%
  \begin{subfigure}[b]{0.32\columnwidth}
  	\centering
  	\includegraphics[width=\textwidth]{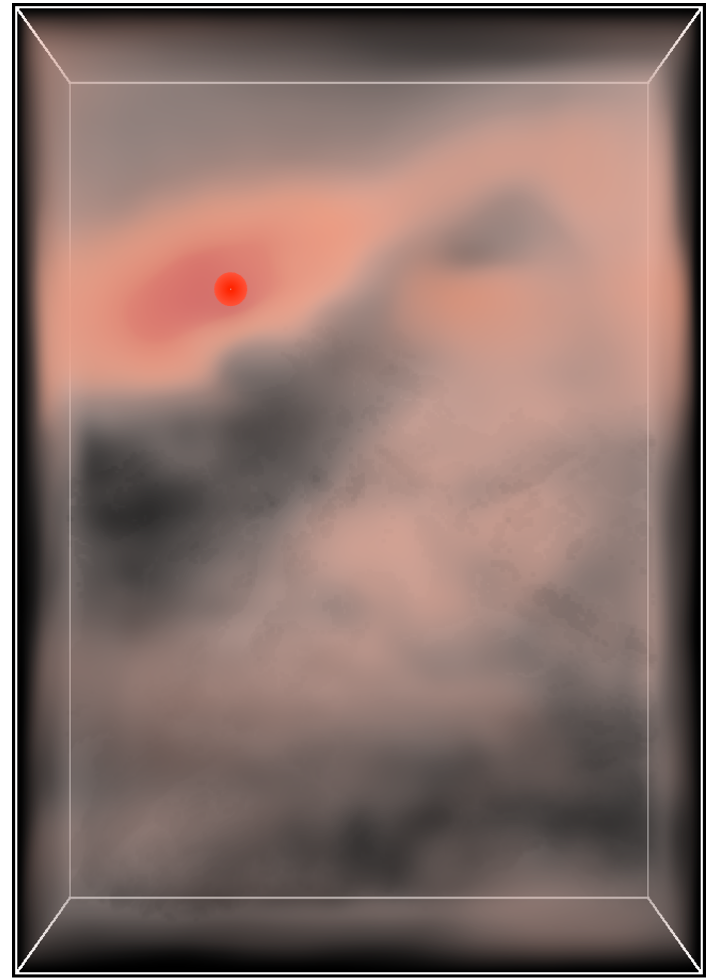}
  	\caption{NDF reconstruction}
  	\label{fig:teaser:mi:ndf}
  \end{subfigure}%
  \begin{subfigure}[b]{0.07\columnwidth}
  	\centering
  	\includegraphics[width=\textwidth]{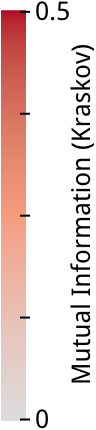}
  	\parbox{\textwidth}{~}
%   	\caption{}
  	\label{fig:teaser:mi:tf}
  \end{subfigure}%
  \caption{
    Neural dependence fields have learned to infer 1500 billion point-to-point Pearson correlation (left) or mutual information estimates (right) in a 1000-member simulation ensemble. Inference of the dependencies between data values at an arbitrary grid vertex (red dot) to all other vertices in a $250 \times 352 \times 20$ grid takes 9\,ms on a high-end GPU. Ground truth volume renderings and network results, respectively, are shown in Figures \ref{fig:teaser:pearson:gt}, \ref{fig:teaser:mi:gt} and \ref{fig:teaser:pearson:ndf}, \ref{fig:teaser:mi:ndf}. The network requires only 1\,GB at runtime.
    %   In a simulation ensemble comprising 1000 members, each discretized on a $250 \times 352 \times 20$ grid, neural network have learned the point-to-point Pearson correlation (left) and estimates of mutual information (right). Reconstructing the learned dependencies between the data values at an arbitrary grid vertex (red dot) to all other vertices takes 9\,ms on a recent high-end GPU. Volume renderings of ground truth fields are in Figures \ref{fig:teaser:pearson:gt}) and \ref{fig:teaser:mi:gt}), the network results in Figures \ref{fig:teaser:pearson:ndf}) and \ref{fig:teaser:mi:ndf}). Only the network requiring 1\,GB must be stored in memory at runtime.
  }
  \label{fig:teaser}
}
\maketitle
%-------------------------------------------------------------------------
\begin{abstract}
We present neural dependence fields (NDFs) -- the first neural network that learns to compactly represent and efficiently reconstruct the statistical dependencies between the values of physical variables at different spatial locations in large 3D simulation ensembles. Going beyond linear dependencies, we consider mutual information as an exemplary measure of non-linear dependence. We demonstrate learning and reconstruction with a large weather forecast ensemble comprising 1000 members, each storing multiple physical variables at a $250 \times 352 \times 20$ simulation grid. By circumventing compute-intensive statistical estimators at runtime, we demonstrate significantly reduced memory and computation requirements for reconstructing the major dependence structures. This enables embedding the estimator into a GPU-accelerated direct volume renderer and interactively visualizing all mutual dependencies for a selected domain point. 


\end{abstract}  
%-------------------------------------------------------------------------
\section{Introduction}

% Estimating statistical dependencies between physical variables at different spatial locations is important to improve the understanding of physical systems in various fields of science and engineering. 
% One example of striking importance is the modeling of atmospheric processes in meteorology, where extensive numerical simulations are carried out to enable accurate forecasting of future weather. As the temporal evolution of the earth system naturally involves many sources of randomness, weather forecasts are required to provide probabilistic statements about the likelihood of future weather events. Operationally, this is achieved by running ensembles of numerical simulations in which slight variations between initial conditions and model specifications account for the uncertainty of initial conditions and model specifications. To translate the resulting volumetric ensemble fields into reliable forecasts, it is important to understand statistical relations, such as spatio-temporal auto-correlations within forecast fields or correlations and coherence between different forecast variables.
Estimating statistical dependencies between physical variables at different spatial locations is crucial for understanding physical systems in various scientific and engineering fields. An important application lies in meteorology, where accurate weather forecasting relies on extensive numerical simulations. Weather forecasts need to account for randomness, and ensembles of simulations with varying initial conditions and model specifications are used to quantify uncertainty. It is essential to analyze statistical relations, such as spatio-temporal auto-correlations within forecast fields or correlations between different forecast variables, to translate volumetric ensemble fields into reliable forecasts.
Studying statistical dependence between random variables is well-researched, and measures exist for assessing linear and non-linear relationships. However, determining relations in 3D ensemble fields presents challenges. Computing dependencies on the fly may be computationally costly, and storing all point-to-point correlations leads to an explosion in required memory. For example, in a simulation ensemble with 1000 members on a $250 \times 352 \times 20$ grid, storing all correlations would need over 3 terabytes of memory, making it infeasible. Additionally, computing correlations between arbitrary point pairs on the fly requires the entire ensemble to fit into the working memory, and more complex measures like mutual information (MI) take roughly 16 minutes on a recent multi-core CPU, hindering interactive analysis of correlation structures.

In this work, we address these challenges by introducing neural dependence fields (NDFs), a novel compact representation of the major correlation structures in large multi-variable ensembles. For this, we propose to interpret fields of two-point correlation measures in 3D ensembles as scalar fields $R$ over the domain of position pairs in 3D space, i.e., $R: \Omega \times \Omega \rightarrow \mathbb{R}$, for $\Omega \subset \mathbb{R}^{3}$. 
Taking inspiration from recent progress in neural scene representations and multi-dimensional tensor decomposition, we design a neural network architecture that exploits self-similarity in the correlation fields to learn a compact representation thereof. At the same time, the network enables fast sampling out of the neural representation. 
Thus, we can avoid holding the ensemble in memory and are able to speed up the computation of correlation estimates significantly, especially for complex non-linear dependence measures. For the ensemble considered in this work, it takes roughly 9\,ms to reconstruct dependencies between an arbitrary reference point and all other points in the domain. 
This allows embedding the network into an interactive volume rendering pipeline, which enables instant visualization and comparison of single-variable auto-correlation and inter-variable correlation fields. In summary, our contributions are:

\begin{itemize}
    \item A compact neural network architecture to learn statistical point-to-point correlations in large ensemble fields. 
    % \item A network-based decoder that makes use of the tiny-cuda-nn framework \cite{tiny-cuda-nn} to enable interactive reconstruction of the dependency structures from the learnt latent-space encoding. \kevin{I would avoid to state the contribution "we use a pre-existing library"}
    \item The embedding of neural network-based correlation reconstruction into direct volume rendering to enable interactive visual exploration of the dependencies in the 3D domain. 
    % \kevin{Does the renderer support isosurfaces? Would be good to know whether we can have a comparison of iso-surface accuracy.} \christoph{Yes, it supports isosurfaces.}
    % \item A comparison of the method against a subsampling-based baseline methods.
    \item A demonstration of interactive correlation analysis for a large meteorological 3D ensemble field.
    %, including spatial auto-correlations of single variables fields and inter-variable correlations. 
\end{itemize}

The proposed method is agnostic towards the choice of correlation measure, such that both linear (e.g., Pearson correlation) and compute-intensive non-linear measures (e.g., MI) are supported. 
The network manages to reconstruct the major correlation structures faithfully, despite showing a tendency to smooth out fine details (cf.\ Fig.~\ref{fig:teaser}). 
% This is even though we have eased the learning process by incorporating symmetry properties into the proposed architecture.
In view of the complexity of the information to be learned, our results demonstrate the potential of network-based correlation learning and open the door for future research in this field, e.g., by looking into more powerful architectures or specialized loss functions. The code for the project is publicly available at \cite{NDFZenodo, CorrerenderZenodo}. %\cite{CorrerenderZenodo}.

%-------------------------------------------------------------------------
\section{Related work\label{sec:relwork}}

\textbf{Scene representation networks and neural fields} Scene representation networks (SRNs) are neural networks trained to derive compact representations of 3D models and scenes. Originally, they were proposed for 2D or 3D position coordinate mapping. Early examples include encoding surface models as implicit functions or occupation maps using fully-connected neural networks \cite{occupancy2019, srn2019chen, park2019deepsdf}. Later, they evolved to encode diverse volumetric scenery information, such as neural radiance fields~\cite{tancik2020,mildenhall2021nerf} and were named neural fields~\cite{xieneural}. A neural field is a neural network that learns a parametrization of spatio-temporal multi-dimensional physical fields over spatial coordinates. In inference, coordinates are transformed into latent-space representations and then decoded to obtain the physical quantity.
Recent work on neural fields has shown that domain-oriented input feature encodings can significantly boost reconstruction quality. Chabra et al. \cite{localsdf2020} proposed laying out trainable parameters in a grid of latent features to learn spatial variations more directly. Refinements include adaptive data structures \cite{martel2021acorn}, fixed multi-resolution grids \cite{takikawa2021}, and multi-resolution spatial hashing \cite{MultiresHashEncoding, muller2021}, enabling multi-scale learning of spatial feature maps. Comprehensive reviews of SRN-related literature focusing on neural scene representations are available \cite{hoang2020, tewari2020nr}.

In scientific data visualization, Lu et al.~\cite{lu2021cnr} introduced SRNs for volumetric data compression, which was sped up and refined by Weiss et al.~\cite{weiss2022fvsrn} through the use of trainable feature representations in combination with an efficient GPU implementation. Höhlein et al.~\cite{hoehlein2022} employ neural fields for compressing ensemble data by sharing model parameters between different ensemble members. Both works demonstrate the combination of volume rendering and network inference as used in this study. 
% Mishra et al.~\cite{mishra2022ftv} use fully-connected neural networks to interpolate scientific data. 
%We aim to build upon and extend these works by focusing explicitly on the capabilities of SRNs in learning dependence fields. 

\noindent \textbf{Correlation visualization} 
% Volume rendering for correlation visualization has been chosen to demonstrate the use of network-based reconstruction in interactive visualization workflows. On the other hand, a number of alternative techniques for correlation visualization have been proposed, e.g., based on clustering \cite{Pfaffelmoser2012,Liebmann2018,Evers2021}, correlation matrices \cite{Chen2011,Evers2021}, diagram views \cite{Sauber2006,Biswas2013,zhangvisual,Liu2016}, or by considering specific features like particle trajectories when analyzing dependencies in flow fields~\cite{Berenjkoub2019}. Correlation sub-sampling ~\cite{gu2010study,Chen2011} identifies the most prominent features in correlation fields and restricts the correlation analysis to those. The aforementioned studies are orthogonal to our contribution in that they either address means to determine significant structures in a given correlation field or develop effective visual encodings of correlation structures. Our approach can be integrated seamlessly into these techniques to reduce memory and computation requirements when accessing correlations in large 3D fields. 
Volume rendering was chosen to demonstrate network-based reconstruction for correlation visualization in interactive workflows. However, alternative techniques for correlation visualization have been proposed, including clustering \cite{Pfaffelmoser2012, Liebmann2018, Evers2021}, correlation matrices \cite{Chen2011, Evers2021}, diagram views \cite{Sauber2006, Biswas2013, zhangvisual, Liu2016}, and specific feature-based approaches like analyzing dependencies in flow fields with particle trajectories \cite{Berenjkoub2019}. Correlation sub-sampling \cite{gu2010study, Chen2011} identifies prominent features in correlation fields for analysis. These approaches complement our contribution, as they address significant structures in correlation fields or develop effective visual encodings. Our approach seamlessly integrates with these techniques, reducing memory and computation requirements for accessing correlations in large 3D fields.
%However, none of these approaches is capable of enabling an interactive visual analysis of the correlation structures in 3D ensemble fields.

%-------------------------------------------------------------------------
\section{Statistical dependence in ensemble fields \label{sec:measures}}

We quantify statistical dependencies in ensemble datasets through bivariate correlation measures $\rho: \mathbb{R}^{N} \times \mathbb{R}^{N} \rightarrow \mathbb{R}$ between vectors of paired random samples. 
For this, let $\Omega\subset \mathbb{R}^{3}$ be a simulation domain in 3D space, 
and let $\mathcal{E} = \{E_{i}: 0 \le i < N\}$ be an ensemble of $N$ multi-variable fields 
$E_{i}: \Omega \rightarrow \mathbb{R}^{d}$. 
The index $i$ suggests a fixed but arbitrary enumeration of the members. For all $i$ and $0 \le \nu < d$, let $E_{i}^{\nu}: \Omega \rightarrow \mathbb{R}$ denote the scalar field associated with variable $\nu$ in member $E_i$. For a given position $\mathbf{p} \in \Omega$, 
we refer to the local sample of variable values as $\mathbf{e}^{\nu}\!(\mathbf{p}) := (E_i^{\nu}(\mathbf{p}): 0\le i \le N) \in \mathbb{R}^{N}$. 
Then, for variables $\mu$ and $\nu$, we consider the field of $\mu$-$\nu$-correlations, 
$R_{\mu\nu}: \Omega \times \Omega \rightarrow \mathbb{R}$, 
where for all pairs of positions $(\mathbf{p}_{\mu}, \mathbf{p}_{\nu}) \in \Omega^2$ the field value is defined as $R_{\mu\nu}\!(\mathbf{p}_{\mu}, \mathbf{p}_{\nu}) := \rho\!(\mathbf{e}^{\mu}\!(\mathbf{p}_{\mu}), \mathbf{e}^{\nu}\!(\mathbf{p}_{\nu}))$. Note that the indexing of the position variables is used to imply that position $\mathbf{p}_\mu$ (position $\mathbf{p}_\nu$) alters the value of $R_{\mu\nu}\!(\cdot,\cdot)$ by changing the reference position in field $\mu$ (field $\nu$), respectively. Special attention is payed to the case $\mu = \nu$, which we refer to as the $\mu$-self-correlation field $S_{\mu} := R_{\mu\mu}$. Notably, $S_{\mu}$ is symmetric under exchange of position coordinates, i.e., $S_{\mu}(\mathbf{p}_{1}, \mathbf{p}_{2}) = S_{\mu}(\mathbf{p}_{2}, \mathbf{p}_{1})$ for all $\mathbf{p}_{1}, \mathbf{p}_{2} \in \Omega$.

% In this work, we use Pearson correlation and MI from the wide class of possible dependence measures. Pearson correlation and MI represent opposite sides of the spectrum regarding computational cost, and, as shown in previous works in visualization~\cite{Berenjkoub2019}, both can indicate different dependence structures and hint at different relationships. 

This work uses Pearson correlation and MI from the wide range of possible dependence measures. Both represent opposite sides of the spectrum of computational cost and indicate different kinds of dependence~\cite{Berenjkoub2019}.
%Thus, by considering both we can effectively assess the overall performance improvements that can be achieved with network-based inference for a wide range of algorithms. 

\subsection{Pearson product-moment correlation coefficient}
% The Pearson Product-Moment Correlation Coefficient, often shortened to Pearson correlation coefficient or Pearson's $r$, is a statistical measure of linear correlation between pairs of random variables. Pearson correlation is often used in data visualization to explore and understand the relationships between different variables.
The Pearson correlation coefficient, or Pearson's $r$, measures the linear correlation between random variable pairs. It is commonly used in data visualization to explore relationships between variables.
Given a set of paired random samples, $\mathbf{e}_1, \mathbf{e}_2 \in \mathbb{R}^N$, the Pearson correlation coefficient is defined as
\begin{equation}
    r = \frac{\cov(\mathbf{e}_1, \mathbf{e}_2)}{\sqrt{\var(\mathbf{e}_1) \var(\mathbf{e}_2)}}\text{,}
\end{equation}
wherein $\var(\cdot)$ and $\cov(\cdot)$ denote sample variance and covariance of the respective random samples. 
With a range of -1 to 1, Pearson correlation indicates correlation (+1), anti-correlation (-1), or the absence of correlation (0). It is easy to interpret, quantifying the strength and direction of the relationship between variables. However, caution is needed when the relationship is nonlinear or when outliers are present, as they can significantly affect the correlation coefficient.
% With a value range of -1 to 1, Pearson correlation is capable of indicating the presence of correlation (+1) and anti-correlation (-1) or the absence of correlation (0).  
% One of the key advantages of the Pearson correlation is that it is easy to interpret and understand. It provides a simple way to quantify the strength and direction of the relationship between two variables. However, it should be used with caution when the relationship between the variables is not linear or when outliers are present in the data, as these can significantly affect the value of the correlation coefficient. 

%-------------------------------------------------------------------------
\subsection{Mutual information}
MI is widely used in machine learning, statistics, and information theory \cite{cover1991entropy} to measure similarity or correlation between random variable pairs. Unlike linear correlation, MI can detect non-linear and non-monotonic dependencies that are not evident in covariance.
% MI is defined as the reduction in uncertainty about one variable when the value of another variable is known. 
Mathematically, this is expressed as
\begin{equation}
    \label{eq:mi}
    I(X_1;X_2) = H(X_1) - H(X_1|X_2)  = H(X_2) - H(X_2|X_1)\text{,}
\end{equation}
wherein $X_1$ and $X_2$ are random variables, $I(X_1;X_2)$ is the MI of $X_1$ and $X_2$, $H(X_1)$ is the entropy of $X_1$, and $H(X_1|X_2)$ is the conditional entropy of $X_1$ given $X_2$. Note that MI is symmetric under the exchange of $X_1$ and $X_2$. 

Estimating MI from finite samples $\mathbf{e}_1$ and $\mathbf{e}_2$ of random variables $X_1$ and $X_2$, i.e.\ computing $I(\mathbf{e}_1, \mathbf{e}_2)$, is computationally expensive. Existing algorithms struggle to scale with large sample sizes~\cite{KraskovMI, KDEMI}. More recent copula-based and neural-network-based variational MI estimators offer better performance in high-dimensional data spaces but still pose computational challenges~\cite{CopulaMI, MINE}. In our study, we compute ground truth MI fields using the nearest-neighbor-based estimator of Kraskov et al.~\cite{KraskovMI}, implemented in parallel. However, detailed performance analysis in section \ref{sec:analysis} shows that estimating MI fields for visualization exceeds the time constraints of interactive data analysis.
% Mutual information neural estimation (MINE) is a recent method that uses neural networks to estimate MI \cite{MINE}. MINE trains a neural network to learn a lower bound on MI, which can be used as an estimate thereof. MINE has been shown to outperform traditional MI estimation methods on several benchmark datasets. Yet, separate networks would be required for all pairs of points in a domain and many network evaluations were needed at run time, which turns out to be inefficient in our scenario.
Mutual information neural estimation (MINE) is a recent method that uses neural networks to estimate MI \cite{MINE}. It trains a neural network to learn a lower bound on MI, which provides an estimate. MINE has shown better performance than traditional MI estimation methods on benchmark datasets. 
Nevertheless, in our scenario, the inefficiency arises because MINE operates on a member-wise basis and requires all members to be present in memory.
% However, in our scenario, it is inefficient to use separate networks for all pairs of points and numerous network evaluations at runtime.

%------------------------------------------------------------------------
\section{Neural dependence fields\label{sec:learning}}

To enable the use of large sets of point-to-point dependence measures, we perform the computationally expensive calculations of these measures in a preprocess and encode the two-point $\mu$-$\nu$-correlation fields $R_{\mu\nu}$ (as defined in section \ref{sec:measures}) with memory- and compute-efficient neural scene representations, $\Phi_{\mu\nu}: \Omega \times \Omega \rightarrow \mathbb{R}$. The network is obtained by solving the optimization problem  
\begin{equation}
    \Phi_{\mu\nu} := \argmin_\Phi \ \mathbb{E}_ {(\mathbf{p}_1, \mathbf{p}_2) \sim \mathcal{U}(\Omega^2)}\left[d\!(\Phi(\mathbf{p}_1, \mathbf{p}_2), R_{\mu\nu}(\mathbf{p}_1, \mathbf{p}_2))\right]\text{,}\label{eq:optim}
\end{equation}
wherein $\Phi\!(\cdot, \cdot)$ is the neural network, $d\!(\cdot, \cdot)$ is a similarity metric, such as $L_1$ or $L_2$ loss, and the expectation $\mathbb{E}\!\left[\cdot\right]$ is taken over samples of position pairs from a uniform distribution, $\mathcal{U}(\Omega^2)$, with support $\Omega^2$. 
The optimization is carried out iteratively using stochastic gradient descent. 
% Using tractably sized batches of position pairs alleviates the need to simultaneously keep excessive amounts of correlation samples in memory. After training, the parameterization of $\Phi_{\mu\nu}$ constitutes a compact encoding of the correlation field, from which samples can be reconstructed rapidly and directly from the compact representation. Note here that similar compaction cannot be achieved using classical compression approaches for multi-dimensional data, such as TThresh~\cite{tthresh} or SZ~\cite{SZ2016}, since such algorithms would require a fixed discretization of the bi-spatial domain $\Omega\times\Omega$, as well as access to the full correlation matrix during compression, which is computationally infeasible.
% Furthermore, once the correlation network has been trained, there is no need to keep the full ensemble dataset in memory. Thus, the approach scales to huge ensemble sizes.
Using tractably sized batches of position pairs simultaneously avoids storing excessive amounts of correlation samples. After training, $\Phi_{\mu\nu}$ is a compact correlation field encoding, enabling rapid sample reconstruction. Classical compression methods like TThresh and SZ cannot achieve similar compaction due to fixed discretization and computational infeasibility. Additionally, once the correlation network is trained, there is no need to keep the entire ensemble dataset in memory, allowing the approach to scale efficiently to large ensemble sizes.

%-------------------------------------------------------------------------
\subsection{Network architecture\label{sec:network}}

\begin{figure}[tbp]
    \centering
    \includegraphics[width=.45\textwidth]{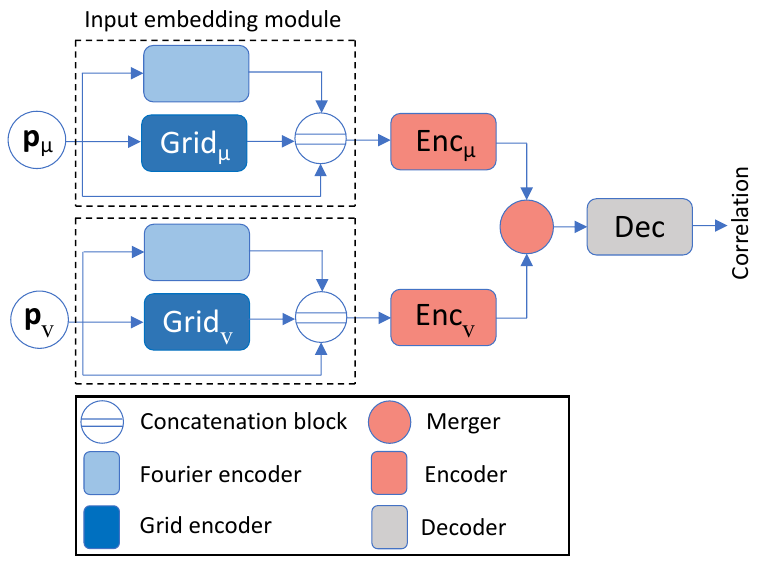}
    \caption{NDF architecture. $\mathbf{p}_{\mu}$ and $\mathbf{p}_{\nu}$ are respectively the reference and query positions. $\grid_\mu$ and $\grid_\nu$, respectively, are the hash grids. Variable-specific encoders $\enc_\mu$ and $\enc_\nu$ are MLPs. Since the feature grids involve trainable parameters, each encoder is equipped with a separate grid. Merger indicates the multiplication of the encoder outputs.
    }
    \label{fig:srn}
\end{figure}

% Achieving accurate reconstruction of two-point correlation fields is non-trivial. In contrast to classical neural scene representations, which work with inputs from 3D spatial or 4D spatio-temporal domains, neural correlation fields operate on a bi-spatial domain $\subset \mathbb{R}^3 \times \mathbb{R}^3$ with variability in a total of six dimensions. This reduces training efficiency since, due to the curse of dimensionality (e.g., \cite{koppen2000curse}), accurately covering the domain with a decent sample density is exponentially more complex. 
% To circumvent generating excessive amounts of correlation samples while at the same time improving the memory efficiency of the network parameterization, we devise NDFs to use sparse sampling information efficiently.
% Achieving accurate reconstruction of two-point correlation fields is challenging. 
NDFs differ from classical neural scene representations as they operate on a bi-spatial domain with six dimensions (6D). Due to the curse of dimensionality, training efficiency is lowered by the additional dimensions since covering the domain adequately with samples becomes exponentially more complex. To overcome this, we construct NDFs to utilize sparse sampling information efficiently, which improves memory efficiency and avoids the computation of excessive amounts of correlation samples. 

% As shown in Figure \ref{fig:srn}, we propose a bipartite network architecture, which is comprised of two variable-specific encoder networks, $\enc_\mu$, and $\enc_\nu$, in combination with a shared decoder network. Both the encoders and decoder are implemented as multi-layer perceptrons (MLPs), composed of $l$ fully-connected layers with $c$ hidden channels and \emph{SnakeAlt} activation~\cite{weiss2022fvsrn}.
% Each encoder model receives information about one of the two positions between which the correlation should be reconstructed as input. Positions are translated into a latent feature vector. These features are merged via element-wise multiplication and forwarded to the decoder, which generates the final prediction. In this way, each encoder is trained on only the marginal space $\Omega\subset\mathbb{R}^3$, with improved training efficiency. The decoder receives preprocessed features, from which the output mapping can be learned more easily.
As shown in Figure \ref{fig:srn}, we propose a bipartite network architecture, which consists of two variable-specific encoder networks, $\enc_\mu$, and $\enc_\nu$, along with a shared decoder network. All networks are implemented as multi-layer perceptrons (MLPs) with $l$ fully-connected layers and $c$ hidden channels using \emph{SnakeAlt} activation~\cite{weiss2022fvsrn}.
Each encoder model receives information about one of the two positions between which the correlation should be reconstructed. Positions are translated into a latent feature vector, which is merged via element-wise multiplication and forwarded to the decoder for the final prediction. This architecture allows each encoder to be trained on only the marginal space $\Omega\subset\mathbb{R}^3$, which improves the training efficiency by increasing the effective amount of correlation samples per volume. In combination with the spatial coherence of the ensemble fields, this enables the model to infer correlations even for point pairs that were not seen during training, thus saving computation time and memory requirements. 
% The decoder receives preprocessed features, making the output mapping easier to learn.

The decomposition is similar in spirit to the approach used in TensoRF~\cite{chen2022tensorf} or K-planes~\cite{fridovich2023k},
% where the authors decompose 3D feature tensors into a linear combination of tensor products between lower-dimensional feature vectors (1D tensor) and matrices (2D tensor) to achieve higher parameter efficiency. In the proposed NDF, a field of features over a 6D domain is decomposed into an outer product of fields over a 3D domain. Note here that prediction accuracy depends crucially on the use of multiplication for feature merging.
where 3D feature tensors are decomposed into linear combinations of tensor products between lower-dimensional feature vectors and matrices for higher parameter efficiency. In the proposed NDF, features over a 6D domain are decomposed into an outer product of fields over a 3D domain. The accuracy of predictions relies heavily on using multiplication for feature merging.
Alternative combination methods, such as concatenation, addition or absolute difference, 
result in a substantial drop in prediction fidelity, which is in line with findings in~\cite{fridovich2023k}. Deviating from TensoRF and K-planes, we found applying MLPs before and after feature merging beneficial, which we validate in more detail in section \ref{sec:analysis}.

For self-correlation fields $S_\mu$ (as defined in section \ref{sec:measures}) and the corresponding NDFs $\Phi_{\mu\mu}$, we further constrain the architecture to use identical encoders for both positions, i.e., $\enc_\mu = \enc_\nu$ (and $\grid_\mu = \grid_\nu$, see below for details). This helps to keep the models small and ensures symmetry of the learned fields under exchange of the query positions on an architectural level, i.e.,\ $\Phi_{\mu\mu}(\mathbf{p}_1, \mathbf{p}_2) = \Phi_{\mu\mu}(\mathbf{p}_2, \mathbf{p}_1)$ is fulfilled trivially by design and does not need to be learned in expensive training iterations.
To improve the capability of the encoders to learn high-frequency patterns as well as spatially distributed and multi-scale features, we employ Fourier features~\cite{mildenhall2021nerf,tancik2020} as well as multi-resolution hash-grids~\cite{MultiresHashEncoding} on the position coordinates, which are concatenated to the raw positions before being processed by the encoders. The input embedding modules are marked with dashed rectangles in Figure \ref{fig:srn}.

\subsection{Input embedding}
For a given vector of input coordinates, $\mathbf{p} = (p_x, p_y, p_z) \in \mathbb{R}^3$, Fourier features increase the spread between spatially close positions by embedding the position information into a higher-dimensional space using the fixed feature mapping
\begin{equation}
    \label{eq:nerf}
    \mathbf{f}_{ij} = (\sin(\omega_i\, \mathbf{n}_j \cdot \mathbf{p}), \cos(\omega_i\, \mathbf{n}_j \cdot \mathbf{p}) )\text{,}
\end{equation}
wherein $\omega_i = 2^i\pi$ for $0 \le i < L \in \mathbb{N}$, and $\mathbf{n}_j\in\mathbb{R}^3$ are the axis-aligned unit vectors for $j\in\{x, y, z\}$. With Fourier features, the model can better resolve high-frequent patterns while not affecting the number of trainable parameters (and thus memory consumption) due to the fixed functional form of the mapping. In our implementation, we empirically determined $L = 12$ as a good choice for the number of Fourier frequencies.

% Multi-resolution hash grids provide a compact means to populate the domain with trainable feature vectors at arbitrary resolution ~\cite{MultiresHashEncoding} using hash tables filled with such feature vectors. By hashing 3D grid indices, the vectors are assigned to regular grid positions at multiple resolution scales, from which feature vectors at arbitrary positions can be retrieved via tri\-linear interpolation. This allows the formation of virtual feature grids at arbitrary resolutions with a fixed memory budget.
Multi-resolution hash grids use hash tables filled with trainable feature vectors to populate the domain at various resolutions~\cite{MultiresHashEncoding}. By hashing 3D grid indices, vectors are assigned to regular grid positions at multiple scales, enabling retrieval of feature vectors at arbitrary positions through tri-linear interpolation. This allows the creation of virtual feature grids at any resolution with a fixed memory budget.

The feature vectors for different resolution levels are concatenated and trained jointly with subsequent model parts using stochastic gradient descent. Hash collisions equilibrate during training due to the pseudo-random hash mapping and multiple resolutions, ensuring adaptive and local feature capacity distribution. The critical parameter for the expressiveness of the hash grid is the hash table size, $2^T$ for $T\in\mathbb{N}$, which also determines memory complexity. Other hyper-parameters include the dimension of the feature vectors per resolution level, the number of resolution levels, and the grid resolution on each level. We use 6 resolution levels with virtual grids of size $16^3$ on the coarsest level, doubling with each finer level. These parameters ensure that the feature granularity matches the spatial resolution of the original dataset, avoiding higher memory consumption with finer levels while maintaining or improving the reconstruction accuracy.
% The multi-resolution hash grids are marked in Figure \ref{fig:srn} with $\grid_\mu$ and $\grid_\nu$, respectively. Since the feature grids involve trainable parameters, the variable-specific encoders $\enc_\mu$ and $\enc_\nu$ are equipped with separate grids.

%-------------------------------------------------------------------------

\subsection{Training}

% During training, we generally consider the case of a rectangular simulation domain $\Omega$ with data samples located on a regular grid. In this setting, we rescale positions such that the simulation domain fills the symmetric unit cube, i.e.,\ $\Omega := [-1, 1]^{3}$. We then generate $10^{6}$ pairs of uniformly distributed random positions $(\mathbf{p}_\mu, \mathbf{p}_\nu)$ in $[-1, 1]^{3}$, use trilinear interpolation to retrieve samples $\mathbf{e}^\mu(\mathbf{p}_\mu)$ and $\mathbf{e}^\nu(\mathbf{p}_\nu)$ from the original ensemble dataset, and compute $R_{\mu\nu}(\mathbf{p}_\mu, \mathbf{p}_\nu)$, as described in section \ref{sec:measures}.
% The models are then optimized according to equation~\ref{eq:optim} using stochastic gradient descent and with $L_1$ loss as a similarity measure. Training with $L_2$ loss instead yielded similar results in our experiments.
During training, we use a rectangular simulation domain $\Omega$ with data samples on a regular grid, rescaled to fill the symmetric unit cube, i.e.,\ $\Omega = [-1, 1]^{3}$. We generate $10^{6}$ pairs of uniformly distributed random positions $(\mathbf{p}_\mu, \mathbf{p}_\nu)$ in $[-1, 1]^{3}$, retrieve samples $\mathbf{e}^\mu\!(\mathbf{p}_\mu)$ and $\mathbf{e}^\nu\!(\mathbf{p}_\nu)$ using trilinear interpolation in the original ensemble dataset, and compute correlations $R_{\mu\nu}(\mathbf{p}_\mu, \mathbf{p}_\nu)$. The models are trained to optimize the $L_1$ loss as a similarity measure, with $L_2$ loss yielding similar results in our experiments.
% We use the Adam optimizer~\cite{adam} with an initial learning rate of $3 \times 10^{-4}$ and 1000 samples per batch. We further employ an adaptive learning rate scheduler, which reduces the learning rate by a factor of $0.1$ if no improvement in reconstruction accuracy is observed for more than 5 passes through the training dataset (epochs). 
We use the Adam optimizer~\cite{adam} with an initial learning rate of $3 \times 10^{-4}$ and 1000 samples per batch. An adaptive learning rate scheduler reduces the learning rate by a factor of $0.1$ after 5 passes without improvement in reconstruction accuracy. After every epoch, the training samples are renewed. The total training duration is 200 epochs.

%-------------------------------------------------------------------------

\section{Correlation visualization\label{sec:vis}}

\begin{figure}[t]
    \centering
    \begin{subfigure}[b]{0.48\columnwidth}
        \centering
      	\includegraphics[width=0.4\textwidth]{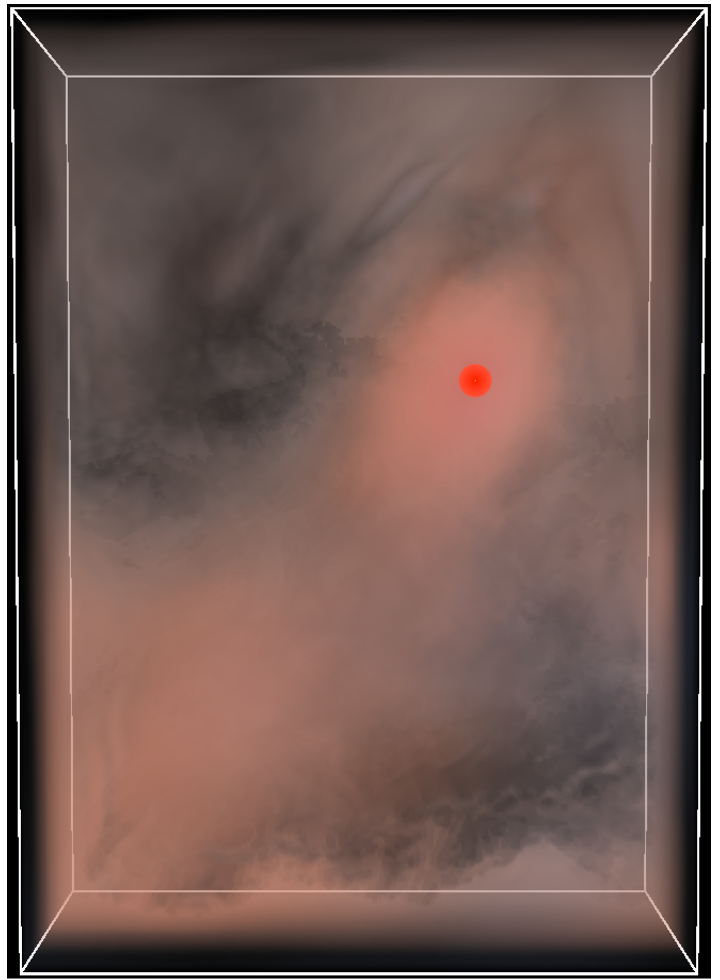}
        \includegraphics[width=0.4\textwidth]{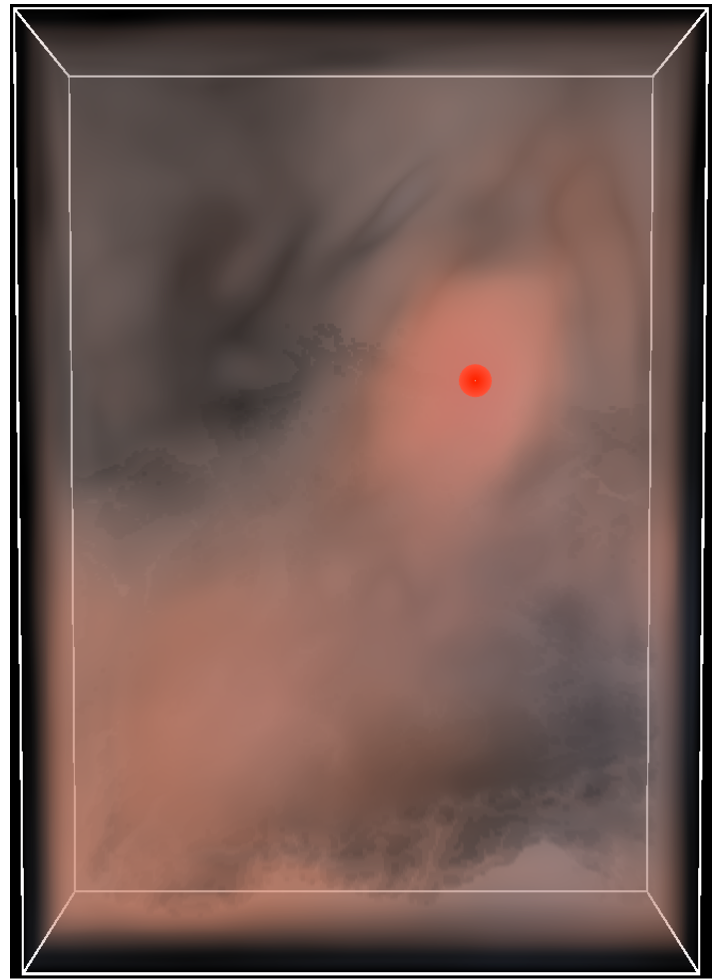}
        \includegraphics[width=0.1\textwidth]{tf/pearson_tf.png}
      	\caption{$\mu = \text{\emph{tk}}$, $\text{PSNR}=87.63$}
    %   	$\mathbf{p}_{\text{ref}} = (0.0, 0.0, 0.0)$ }
    \end{subfigure}% 
    % \begin{subfigure}[b]{0.48\columnwidth}
    %   	\centering
    %   	\includegraphics[width=0.4\textwidth]{pearson/pearson_tk_157_127_10_gt.png}
    %   	\includegraphics[width=0.4\textwidth]{pearson/pearson_tk_157_127_10_srn.png}
    %     \includegraphics[width=0.1\textwidth]{tf/pearson_tf.png}
    %   	\caption{$\mu = \text{\emph{tk}}$, $\text{PSNR}=82.82$} 
    % %   	$\mathbf{p}_{\text{ref}} = (0.256, -0.278, 0.0)$}
    % \end{subfigure}\\
        \begin{subfigure}[b]{0.48\columnwidth}
      	\centering
      	\includegraphics[width=0.4\columnwidth]{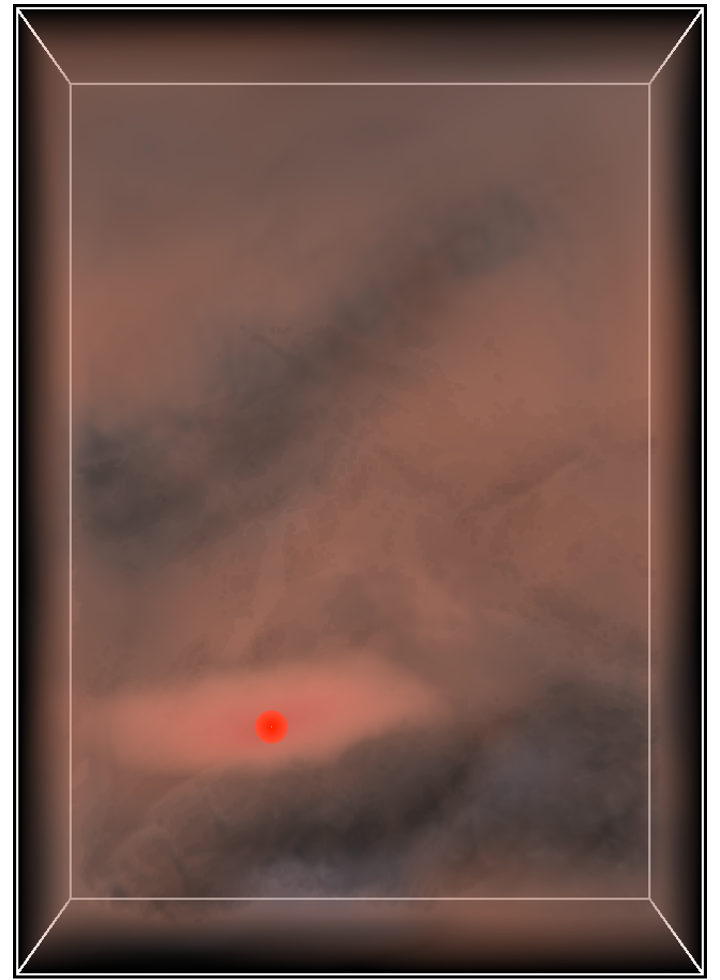}
        \includegraphics[width=0.4\columnwidth]{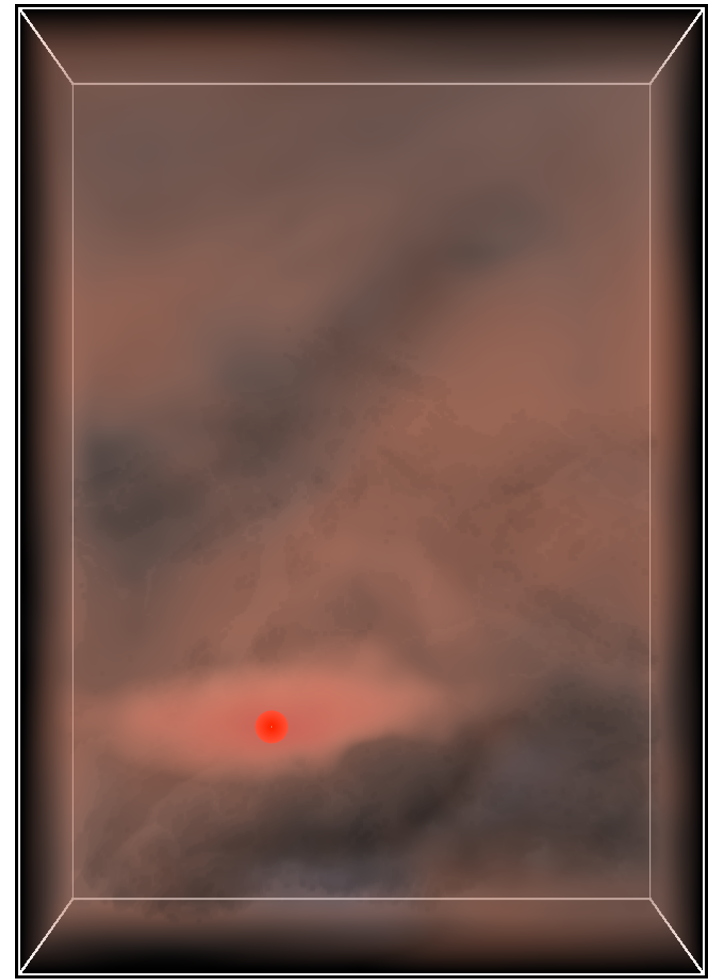}
        \includegraphics[width=0.1\textwidth]{tf/pearson_tf.png}
      	\caption{$\mu = \text{\emph{u}}$, $\text{PSNR}=86.39$}
    %   	$\mathbf{p}_{\text{ref}} = (0.0, 0.0, 0.0)$ }
      \end{subfigure}%
    % \hspace{0.1cm}
    % \begin{subfigure}[b]{0.48\columnwidth}
    %   	\centering
    %   	\includegraphics[width=0.4\columnwidth]{pearson/pearson_u_157_127_10_gt.png}
    %     \includegraphics[width=0.4\columnwidth]{pearson/pearson_u_157_127_10_srn.png}
    %     \includegraphics[width=0.1\textwidth]{tf/pearson_tf.png}
    %   	\caption{$\mu = \text{\emph{u}}$, $\text{PSNR}=86.76$}
    % %   	$\mathbf{p}_{\text{ref}} = (0.256, -0.278, 0.0)$ }
    %   \end{subfigure}%
    \caption{Point-to-point Pearson self-correlations $S_\mu$ for different variables $\mu$ (temperature \emph{tk} and longitudinal component \emph{u} of wind speed) and reference positions $\mathbf{p}_{\text{ref}}$. Figures show ground truth (left) and NDF reconstruction (right). The reference is shown in red.}
        % \caption{Pearson correlation visualization for different points in parameter \emph{tk}. a) Point location $(x=125, y=176, z=10)$, ground truth (left), model reconstruction (right). b) Point location $(x=157, y=127, z=10)$, ground truth (left), model reconstruction (right).}
    \label{fig:Pearson_single_ref}
\end{figure}

% Once trained, the network can infer dependence estimates for arbitrary pairs of positions, and multiple queries can be batched efficiently to foster parallel processing on GPU hardware. 
% To accelerate the reconstruction of learned dependence measures, we employ the tiny-cuda-nn framework~\cite{tiny-cuda-nn}, which offers a fully-fused MLP implementation~\cite{FusedMLP} using fast 16-bit inference with tensor cores on NVIDIA GPUs in combination with multi-resolution hashed feature grids~\cite{MultiresHashEncoding}. The fully-fused MLP implementation in tiny-cuda-nn exploits that for small networks, the weight matrices, which normally reside in global GPU memory, can fit into registers, while intermediate outputs of one layer may fit into shared memory. MLP execution can thus be fused into a single kernel, significantly reducing access operations to global GPU memory and eliminating the overhead of multiple compute kernel invocations per layer of the MLP.
% To support our model architecture, custom activation function like \emph{Snake} and \emph{SnakeAlt}~\cite{weiss2022fvsrn} were added to a fork of the library, as the simpler activation functions like ReLU led to inferior reconstruction accuracy.
Once trained, the network can estimate dependencies for any position pair. Multiple queries can be batched efficiently for parallel processing on a GPU using the tiny-cuda-nn framework~\cite{tiny-cuda-nn}, which features a fully-fused MLP implementation~\cite{FusedMLP} with fast 16-bit inference using tensor cores on NVIDIA GPUs and provides functionality for the multi-resolution hashed feature grids~\cite{MultiresHashEncoding}. The custom activation functions \emph{Snake} and \emph{SnakeAlt}~\cite{weiss2022fvsrn} were added to a fork of the library to support the model architecture, as simpler activation functions like ReLU led to inferior reconstruction accuracy.

Network training is performed in PyTorch using the Python bindings of the library. To enable interactive visualizations, the binary weights of the MLP encoder and decoder can then be loaded by a tiny-cuda-nn module into whatever GPU correlation visualization is used. In our primary use case, we access the network from a GPU-based volume renderer implemented in Vulkan~\cite{VulkanSpec}. The renderer is tied to a graphical interface, in which the user is able to select reference points $\mathbf{p}_{\text{ref}} \in [-1, 1]^3$, for which correlation samples are reconstructed and displayed as a density field. Specifically, we consider the case of displaying volumetric correlation fields, $\phi: [-1, 1]^3 \rightarrow \mathbb{R}$, where $\phi(\mathbf{p}) := \Phi_{\mu\nu}(\mathbf{p}, \mathbf{p}_{\text{ref}})$ or $\Phi_{\mu\nu}(\mathbf{p}_{\text{ref}}, \mathbf{p})$. 

% The correlation fields are sampled on a regular volumetric grid with resolution $X\times Y \times Z$ for visualization. Upon selection of $\mathbf{p}_{\text{ref}}$, the volume renderer prepares a CUDA input buffer for the reference point. The buffer is then passed to the tiny-cuda-nn encoder module to get the encoded reference vector. Then, the grid is subdivided into $\lceil (X \times Y \times Z)/M \rceil$ query batches of size $M$. For each batch, a subset of grid positions is given to the encoder, which outputs the set of encoded feature vectors for the grid subset. Finally, the reference and query features are multiplied and passed to the tiny-cuda-nn decoder module, which outputs a set of correlation estimates to a buffer. The output buffer is a Vulkan buffer shared with CUDA via Vulkan-CUDA interoperability features. Finally, the content of the shared output buffer is copied to a 3D Vulkan image for visualization in the volume renderer. Race conditions are avoided by using semaphores shared between CUDA and Vulkan. 
For visualization, the correlation fields are sampled on a grid with resolution $X\times Y \times Z$. A CUDA input buffer is prepared for the reference point $\mathbf{p}_{\text{ref}}$ and passed to the tiny-cuda-nn encoder module to get the encoded reference vector. The grid is divided into $\lceil (X \times Y \times Z)/M \rceil$ query batches of size $M$. Each batch is encoded, and the reference and query features are multiplied before being passed to the tiny-cuda-nn decoder module to obtain correlation estimates in an output buffer shared between Vulkan and CUDA, preventing race conditions with shared semaphores. Finally, the shared output buffer is copied to a 3D Vulkan image for visualization in the volume renderer.

% In general, the aforementioned implementation can also be used for reconstructing correlation samples in turn during ray marching. In this case, instead of issuing the positions of grid vertices to the network, all sample positions at the $i$-th ray-marching step are issued and reconstructed. Note that the timings we give below refer to the first option, i.e., the correlation grid is first reconstructed and ray-marching is performed on this grid. Since network inference is embedded into the volume ray-caster, both direct and iso-surface volume rendering can be performed. Visualizations are discussed in more detail in section \ref{sec:addexp}
%-------------------------------------------------------------------------
\section{Performance and Quality Analysis\label{sec:analysis}}

% In the following, we demonstrate the use of the NDF model for interactive visual analysis of the spatial statistical dependencies between the same or multiple variables in a large weather forecast ensemble. We shed light on the reconstruction speed of the network as well as the memory requirements of the network architecture. The results are qualitatively compared to ground truth dependence fields, which have been computed using floating point implementations of Pearson correlation coefficients on the GPU and the CPU and MI on the CPU. 
Here, we showcase the NDF model for interactive visual analysis of spatial statistical dependencies in a large weather forecast ensemble. We examine the network's reconstruction speed and memory requirements and compare the results to ground truth dependence fields obtained using Pearson correlation coefficients and MI on GPU and CPU.

\begin{figure}[t]
    \centering
    \begin{subfigure}[b]{0.48\columnwidth}
      	\centering
      	\includegraphics[width=0.4\textwidth]{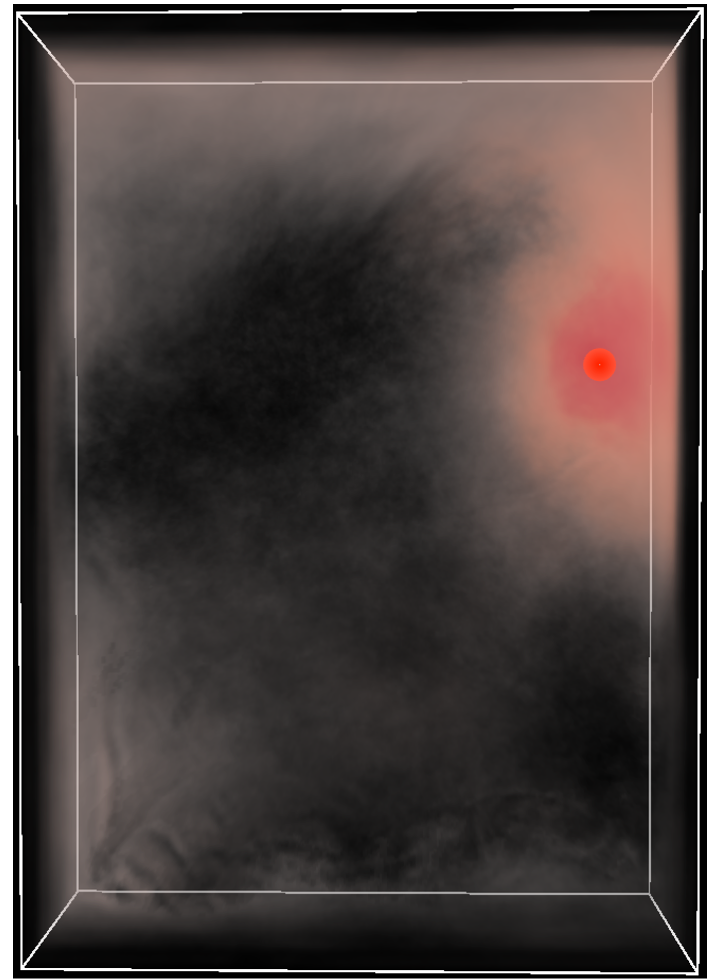}
        \includegraphics[width=0.4\textwidth]{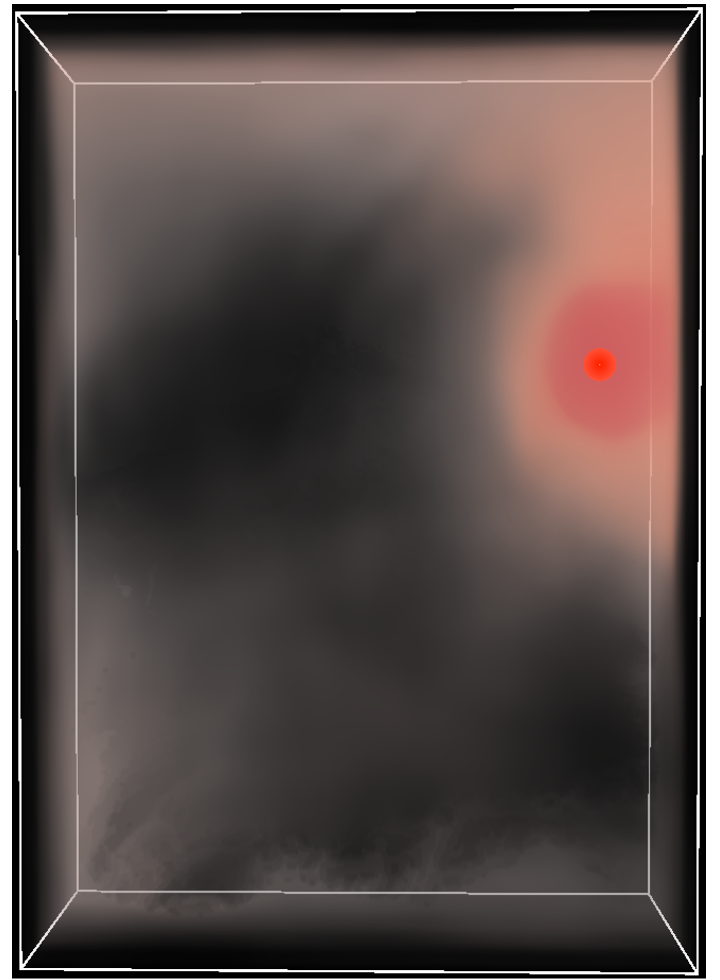}
        \includegraphics[width=0.1\textwidth]{tf/mi_tf_05.png}
      	\caption{$\mu = \text{\emph{tk}}$, $\text{PSNR}=86.44$}
    %   	$\mathbf{p}_{\text{ref}} = (0.136, -0.290, 0.0)$ }
    \end{subfigure}%
    % \begin{subfigure}[b]{0.48\columnwidth}
    %   	\centering
    %   	\includegraphics[width=0.4\textwidth]{/mi/mi_tk_210-320-10_gt_08.png}
    % \includegraphics[width=0.4\textwidth]{/mi/mi_tk_210-320-10_srn_08.png}
    % \includegraphics[width=0.1\textwidth]{tf/mi_tf_05.png}
    %   	\caption{$\mu = \text{\emph{tk}}$, $\text{PSNR}=83.64$}
    % %   	$\mathbf{p}_{\text{ref}} = (0.680, 0.818, 0.0)$ }
    % \end{subfigure}\\
        \begin{subfigure}[b]{0.48\columnwidth}
      	\centering
      	\includegraphics[width=0.4\textwidth]{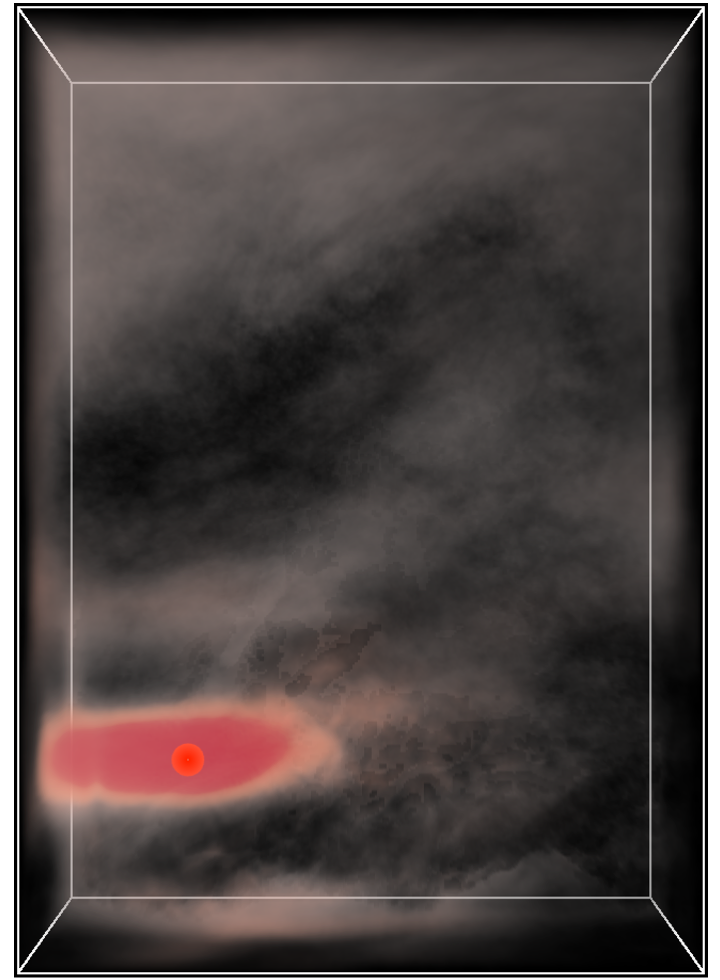}
        \includegraphics[width=0.4\textwidth]{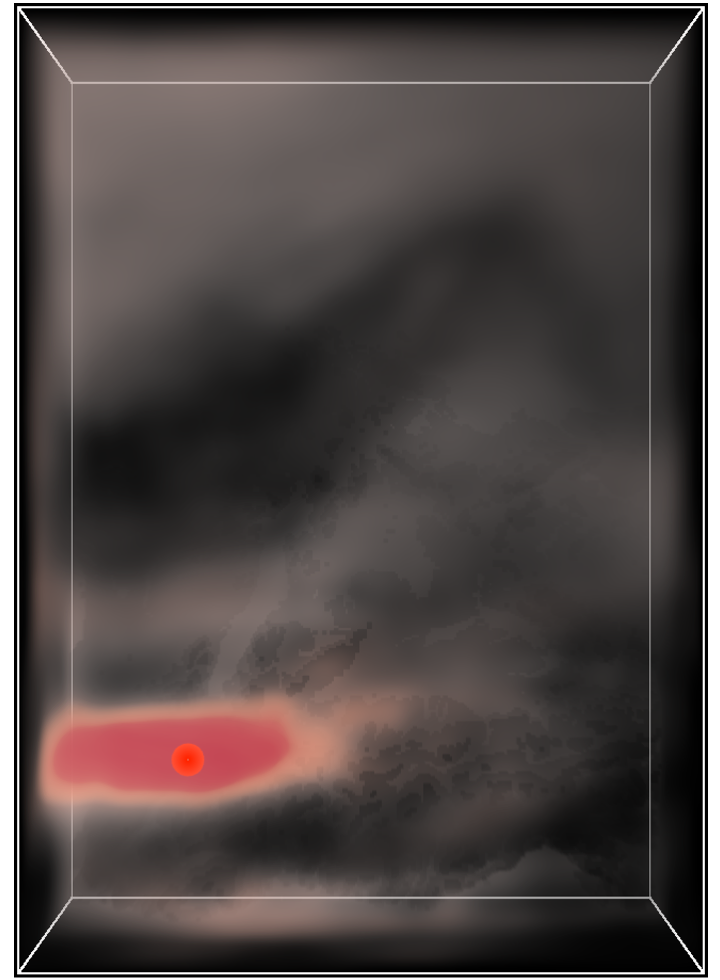}
        \includegraphics[width=0.1\textwidth]{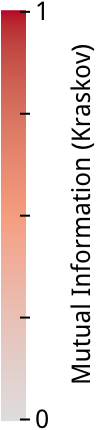}
      	\caption{$\mu = \text{\emph{u}}$, $\text{PSNR}=85.27$}
    %   	$\mathbf{p}_{\text{ref}} = (0.0, 0.0, 0.0)$ }
    \end{subfigure}%
    % \begin{subfigure}[b]{0.48\columnwidth}
    %   	\centering
    %   	\includegraphics[width=0.4\textwidth]{mi/mi_u_157_127_10_gt.png}
    % \includegraphics[width=0.4\textwidth]{mi/mi_u_157_127_10_srn.png}
    % \includegraphics[width=0.1\textwidth]{mi/mi_u_157_127_10_tf.png}
    %   	\caption{$\mu = \text{\emph{u}}$, $\text{PSNR}=83.73$}
    % %   	$\mathbf{p}_{\text{ref}} = (0.256, -0.278, 0.0)$ }
    % \end{subfigure}%
    \caption{Point-to-point MI self-correlations $S_\mu$ for different variables $\mu$ (temperature \emph{tk} and longitudinal component \emph{u} of wind speed) and reference positions $\mathbf{p}_{\text{ref}}$.  Figures show ground truth (left) and NDF reconstruction (right). The reference is shown in red.}
    \label{fig:MI_single_ref}
\end{figure}
%-------------------------------------------------------------------------

\subsection{Dataset}

% We validate our approach using a convective-scale multi-variable ensemble dataset (CSEns) created by Necker et al.~\cite{Necker2020}. The ensemble comprises 1000 numerical simulations of a 3D atmospheric dynamics model. The simulation domain is a rectangular region in central Europe, covered by a regular grid with $250 \times 352$ nodes. 
% Vertical variations are captured on 20 discrete levels in height, and the simulations span six hours in total. Following Höhlein et al.~\cite{hoehlein2022}, we select the time step with the largest simulation time, which, upon visual inspection, displays the most interesting features. The generalization of our approach to multiple time steps and inter-temporal dependencies is subject to future work and discussed in section \ref{sec:conclusion}.
% The CSEns dataset contains 3D data for nine different meteorological variables. In this work, we show visualizations for temperature (tk) and longitudinal wind (u), which serve well in illustrating representative findings.
% We validate our models by computing ground truth correlation fields for the selected time step and variables using the complete set of 1000 ensemble members.
We validate our approach using a convective-scale multi-variable ensemble dataset (CSEns) by Necker et al.~\cite{Necker2020}. It consists of 1000 numerical simulations of a 3D atmospheric dynamics model over a rectangular region in central Europe, with a grid size of $250 \times 352$ nodes and 20 discrete height levels. The simulations span six hours, and we select the last time step with the most interesting features for visualization~\cite{hoehlein2022}. The dataset includes 3D data for nine meteorological variables. For validation, we compute ground truth correlation fields for temperature (tk) and longitudinal wind (u) using all 1000 ensemble members. The generalization of our approach to multiple time steps and inter-temporal dependencies will be explored in future work.

\subsection{NDF model performance}

\begin{table}[b]
\caption{Performance comparison. For a selected grid point, the dependence measures are computed for all other grid points in a $250 \times 352 \times 20$ simulation grid using a single variable. A GPU implementation of the MI estimator is not available.    \label{tab:performance}}
\begin{center}
    \renewcommand{\arraystretch}{1.25}
    \begin{tabular}{c|c|c|c}
    %   \hline\hline
      \ & \textbf{NDF (ours)} &  \textbf{Pearson} & \textbf{MI \cite{KraskovMI}}\\
    %   \hhline{=|=|=|=}
      \hline \hline
      CPU & -- & $4772\,\text{ms}$ & $1026957\,\text{ms}$ \\
      \hline
      GPU & $9\,\text{ms}$ & $234\,\text{ms}$ & -- \\
      \end{tabular}
\end{center}
\end{table}

To shed light on the performance of the NDFs, we conducted a comparative analysis of the runtime of NDFs against reference implementations of Pearson correlation and MI. The implementation of the MI estimator follows Kraskov et al.~\cite{KraskovMI}. All performance measurements are based on the CSEns dataset and were performed on an NVIDIA RTX 3090 GPU and a 6-core Intel Xeon W-2235 CPU, respectively. Notably, a parallel MI estimator is only available on the CPU, whereas we restricted ourselves to a GPU implementation of NDF-based reconstruction. Table \ref{tab:performance} shows that our model is about 26x faster than the GPU implementation of the Pearson correlation coefficient.
%-- the computationally most simple dependence measure. 
Compared to the CPU implementation of the MI estimator, the factor is $114{,}106\times$. Once the NDF model is trained, it takes 9\,ms to reconstruct the dependencies between the data values at a selected grid point and all other grid points. The network model requires roughly 1\,GB of memory while keeping the entire dataset in memory for one variable would amount to 7\,GB. Training of the NDF takes approximately one hour on the aforementioned machine. Even though it is difficult to estimate the performance of a GPU implementation of the MI estimator, it can be assumed that even an optimized GPU-accelerated estimator will be significantly slower than the NDF model. The MI estimator must construct search structures for nearest-neighbour queries for all requested samples, a much more elaborate process than passing data through fully-fused MLP kernels. 
%-------------------------------------------------------------------------
\subsection{NDF model accuracy}

\begin{figure}[t]
    \centering
    \includegraphics[width=\linewidth]{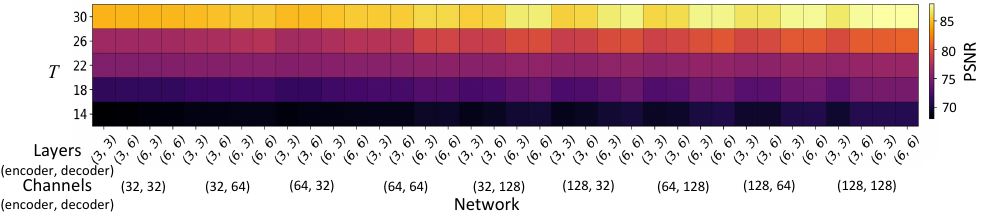}
    \caption{Impact of hash table size and MLP hyperparameters on NDFs' reconstruction quality. The horizontal axis displays various encoder and decoder configurations, including the number of layers and hidden channels. The vertical axis shows the log-2 hash table size $T$ from section \ref{sec:network}. The plot considers variable temperature (\emph{tk}) and Pearson correlation, with similar behavior observed for other variables and similarity metrics.
    % Impact of hash table size and MLP hyperparameters on reconstruction quality of NDFs. The horizontal axis shows different configurations for the encoder and decoder. The number of layers for the encoder and decoder, as well as the number of hidden channels, are respectively shown in two rows in this axis. The vertical axis shows the log-2 hash table size $T$ as introduced in section \ref{sec:network}. Variable temperature (\emph{tk}) and the Pearson correlation are considered in this plot. Other variable and similarity metric show similar behavior.
    }
    \label{fig:ndf_performance_heatmap}
\end{figure}
%The quality of network-based reconstruction is analyzed by comparing the results to the ground truth self-correlation fields $S_\mu$ for Pearson correlation and MI. 
%This characteristics is observed in direct volume renderings, as provided in Figure \ref{fig:teaser}, Figure \ref{fig:Pearson_single_ref} and Figure \ref{fig:MI_single_ref}.
%as well as in iso-surface visualizations, shown in Figure \ref{fig:dvr-iso}. 
%The lack of detail can be due to multiple reasons. 

%\subsection{Input embedding}

% A visual comparison as provided in Figures \ref{fig:teaser}, \ref{fig:Pearson_single_ref} and \ref{fig:MI_single_ref} shows that the network is able to reconstruct the major dependence structures in the variable fields, while fine details cannot be reproduced equally well. One reason might be that model training is restricted to using a comparatively small amount of correlation samples, such that not all details of the complete 6D correlation field can be captured in the training dataset. 
% Thus, the model needs to interpolate into the unseen regions between samples. 
% However, as discussed in section \ref{sec:network}, the proposed bi-partite NDF architecture helps to use sample information more efficiently by increasing the effective sampling density for the encoder parts of the model. 
Figures \ref{fig:teaser}, \ref{fig:Pearson_single_ref}, and \ref{fig:MI_single_ref} visually compare the network's ability to reconstruct major dependence structures in the variable fields. While fine details may not be equally well reproduced due to the limited number of correlation samples during model training, the proposed bi-partite NDF architecture uses sample information efficiently by increasing the effective sampling density for the encoder parts of the model (see section~\ref{sec:network}).
% Furthermore, experiments with an increased number of samples per epoch and with higher resampling frequencies did not improve the reconstruction significantly, from which we conclude that sampling density is not a limiting factor of reconstruction accuracy. This is backed further by observing a trade-off between model capacity, i.e.,\ model memory size, and reconstruction quality. This indicates that the reconstruction quality is primarily bounded through the model's ability to store the training information rather than through sample availability. It needs to be considered here that the shown figures display dependence fields for single selected reference points, i.e.,\ 3D slices into an exponentially larger 6D space of correlations, on which the model has been trained. Theoretically, storing all possible two-point correlations in 32-bit floating point format would amount to 6\,TB of memory space, assuming the data to reside on the CSEns grid of size $250\times352\times 20$. In light of this, a resulting network size of 1\,GB corresponds to an effective compression factor of more than $6,000\times$, which makes the obtained results seem fairly promising. 
Furthermore, experiments with increased sample counts did not significantly improve reconstruction, indicating that sampling density is not the limiting factor for reconstruction accuracy. A trade-off between model capacity (memory size) and reconstruction quality is observed, suggesting that the quality is primarily bounded by the model's ability to store training information rather than sample availability. The figures displayed dependence fields for selected reference points, representing 3D slices in a much larger 6D correlation space where the model was trained. Storing all possible two-point correlations in 32-bit floating point format would require 6\,TB of memory space for the CSEns grid of size $250\times352\times20$. The resulting network size of 1\,GB corresponds to an effective compression factor of over $6{,}000\times$, showing promising results.
% Figure \ref{fig:ndf_performance_heatmap} illustrates the trade-off between model size and reconstruction quality. For Pearson self-correlation fields of variable \emph{tk}, we train NDFs with different settings for the hash-table size $2^T$ and different complexities of encoder and decoder MLPs. PSNR values are computed on a set of $10^6$ position pairs, in which each of the positions is sampled uniformly from the grid domain. For efficiency, we train models for this test shorter, only for 50 epochs.

Figure \ref{fig:ndf_performance_heatmap} shows the trade-off between model size and reconstruction quality. NDFs are trained with various settings for the hash-table size $2^T$ and different complexities of encoder and decoder MLPs, using Pearson self-correlation fields of variable \emph{tk}. PSNR values are computed on a set of $10^6$ position pairs, sampled uniformly from the grid domain. Models are trained for a shorter duration of 50 epochs for efficiency.
% The figure suggests a minor advantage for models with more complex encoder and decoder MLPs. However, the most significant determinant of achievable PSNR is the hash table size, which also has the strongest effect on the model memory consumption. We observe a clear trend of increasing PSNR with increasing Table size. Note here that a doubling of the table size $2^T$ corresponds to roughly a doubling of the model memory requirements, whereas the volume of the MLP parameterization is limited to only a few kB due to restrictions of GPU kernel memory and is thus negligible. In the current implementation of tiny-cuda-nn, we find the maximum table size bounded to less than $2^{32}$. For further experiments, we choose $T = 30$ in combination with 6-layer encoders and decoders with 128 channels per layer each.
The figure indicates a minor advantage for models with more complex encoder and decoder MLPs. However, the most significant factor affecting achievable PSNR is the hash table size, which also strongly affects model memory consumption. Increasing the table size leads to higher PSNR values. A doubling of the table size $2^T$ roughly doubles the model memory requirements, while the volume of the MLP parameterization is limited to only a few kB, making it negligible. In our implementation, we set the maximum table size to less than $2^{32}$. For further experiments, we choose $T = 30$ with 6-layer encoders and decoders, each having 128 channels per layer.

\begin{figure}[t]

    \centering
  	\begin{subfigure}[b]{0.3\columnwidth}
        \centering
      	\includegraphics[width=.9\textwidth]{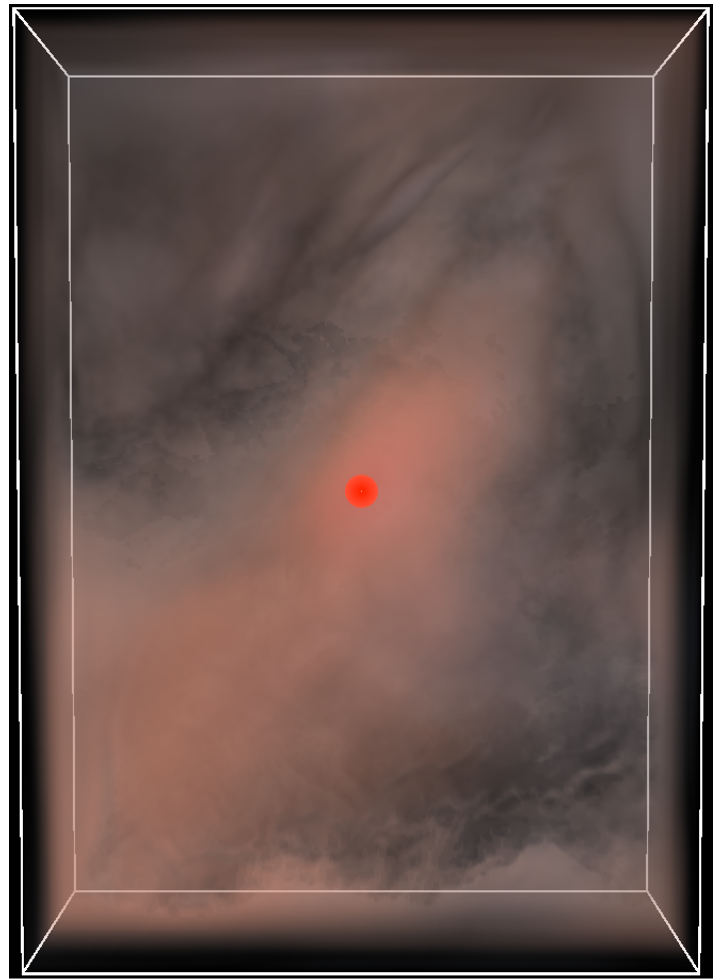}
      	\caption{Ground truth}
    
    \end{subfigure}%
    \begin{subfigure}[b]{0.3\columnwidth}
        \centering
      	\includegraphics[width=.9\textwidth]{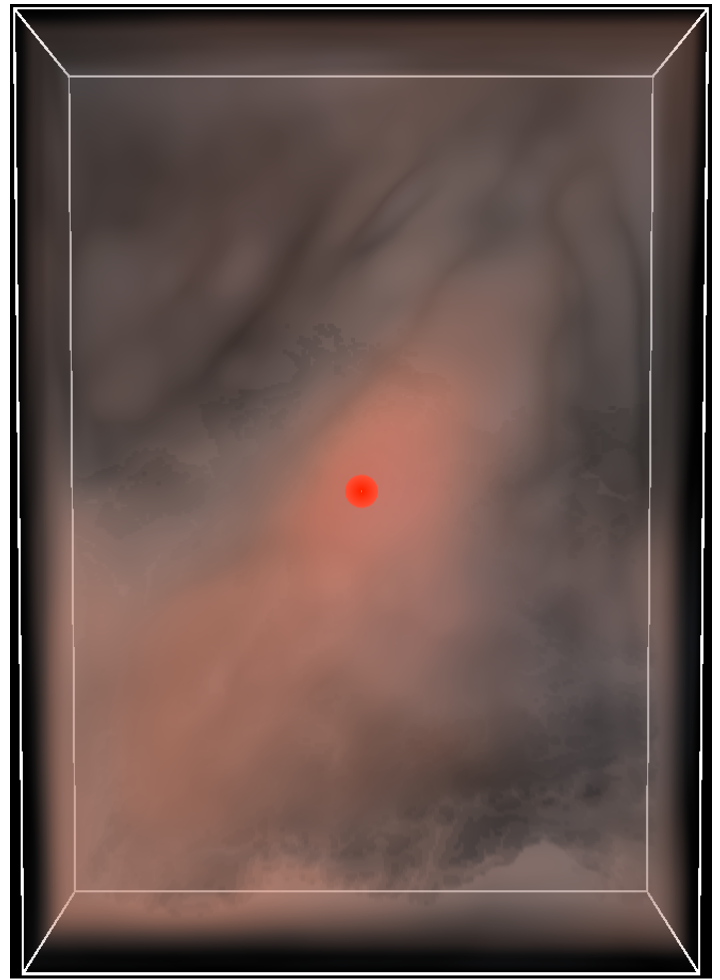}
      	\caption{NDF (ours)}
   
    \end{subfigure}%
    \begin{subfigure}[b]{0.3\columnwidth}
        \centering
      	\includegraphics[width=.9\textwidth]{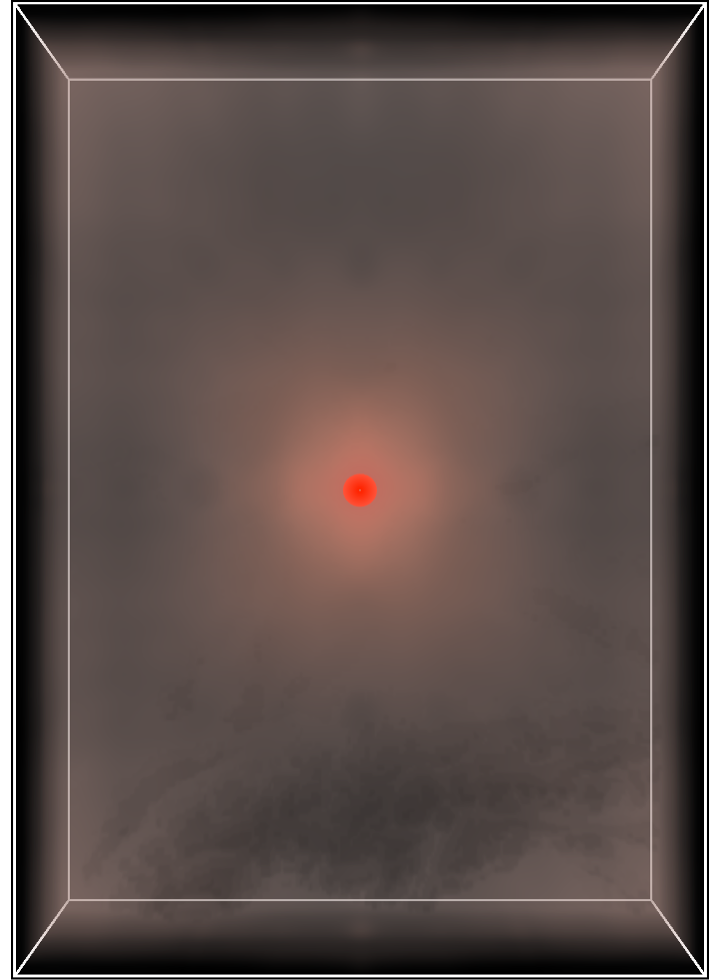}
      	\caption{NDF, grid only}
    
    \end{subfigure}%
    \begin{subfigure}[b]{0.045\columnwidth}
  	\centering
  	\includegraphics[width=\textwidth]{tf/pearson_tf.png}
  	\parbox{\textwidth}{~}
%   	\caption{}
%   	\label{fig:}
  \end{subfigure}%
    \caption{Accuracy comparison between NDFs with MLP encoder (complete) and pure grid model (without encoder) using Pearson self-correlation fields for variable temperature (\emph{tk}).}
    \label{fig:NDF_vs_tensorf}
\end{figure}

% To validate our design choices against TensoRF \cite{chen2022tensorf} and K-planes \cite{fridovich2023k}, we compare the reconstruction quality of the proposed architecture against a TensoRF-like model, in which the MLP in the encoder part is omitted, i.e., \ predictions are based on only the outputs of the input embedding. Figure \ref{fig:NDF_vs_tensorf} displays reconstructed Pearson correlation fields for both approaches and demonstrates the significant added value of using the MLP before feature merging. 
To validate our design choices against TensoRF \cite{chen2022tensorf} and K-planes \cite{fridovich2023k}, we compare the reconstruction quality of the proposed architecture against a TensoRF-like model, where the MLP in the encoder part is omitted, predicting based solely on the outputs of the input embedding. Figure \ref{fig:NDF_vs_tensorf} displays reconstructed Pearson correlation fields for both approaches, highlighting the significant added value of using the MLP before feature merging.

%-------------------------------------------------------------------------

\subsection{Additional experiments\label{sec:addexp}}

% In the following, a number of quantitative results using NDFs for the reconstruction of Pearson correlation coefficients and MI values are shown. We start by conducting experiments using NDFs for variable self-correlation fields $S_\mu$ and corresponding networks $\Phi_{\mu\mu}$. 
The following are quantitative results using NDFs for reconstructing Pearson correlation coefficients and MI values. We begin with experiments on variable self-correlation fields $S_\mu$ and their corresponding networks $\Phi_{\mu\mu}$.

% Firstly, the user selects a specific point in the 3D domain, and instantaneously the dependence field $\phi(\mathbf{r})$, as described in section \ref{sec:vis}, is displayed via volume rendering (as demonstrated in the accompanying video). Volumetric field visualizations can be generated in less than $10\,\text{ms}$, including reconstruction and rendering. Different regions appear highlighted by interactively moving the reference point in the 3D domain to indicate a high correlation against the reference point. In this way, the user can discover islands of high internal correlation or assess the decay of correlations with increasing distance from the reference. To improve visibility of certain features, the transfer function can be adapted interactively with respect to color and opacity throughout the analysis process.
\noindent \textbf{Single-point experiment} Firstly, the user selects a point in the 3D domain, and the dependence field $\phi(\mathbf{r})$ is instantly displayed via volume rendering (see supplementary video). Volumetric visualizations, including reconstruction and rendering, can be generated below 10\,ms. Moving the reference point interactively highlights different regions with high correlation. This allows the user to identify areas of high internal correlation or observe how correlations change with distance from the reference. The transfer function can be adjusted interactively for better visibility of specific features during the analysis.

\begin{figure}
    \centering
    \begin{subfigure}[b]{0.48\columnwidth}
      	\centering
      	\includegraphics[width=0.4\textwidth]{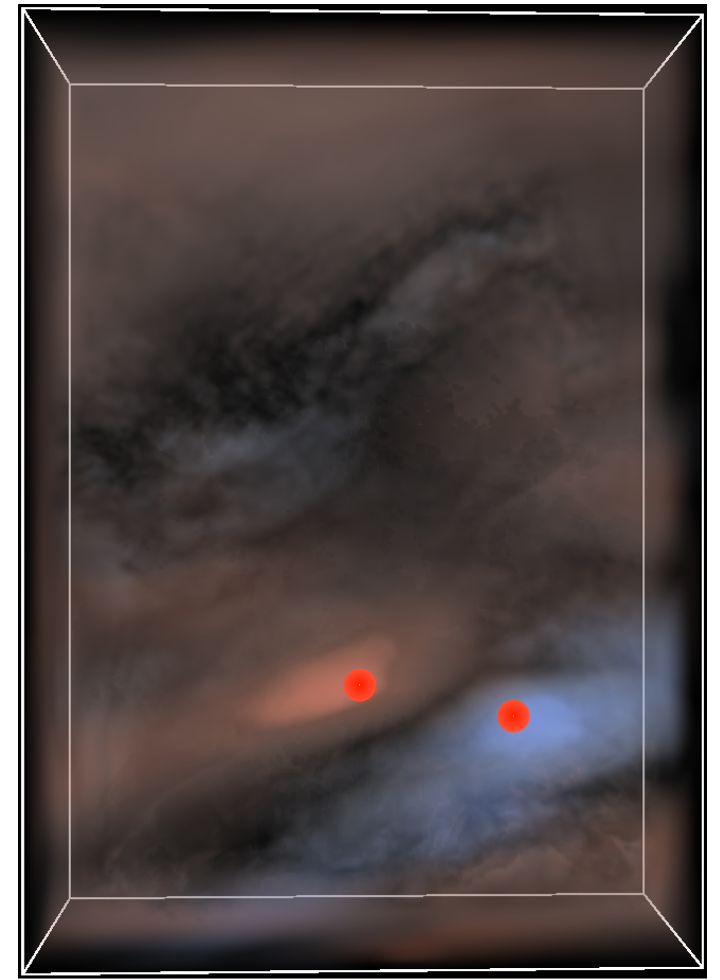}
        \includegraphics[width=0.4\textwidth]{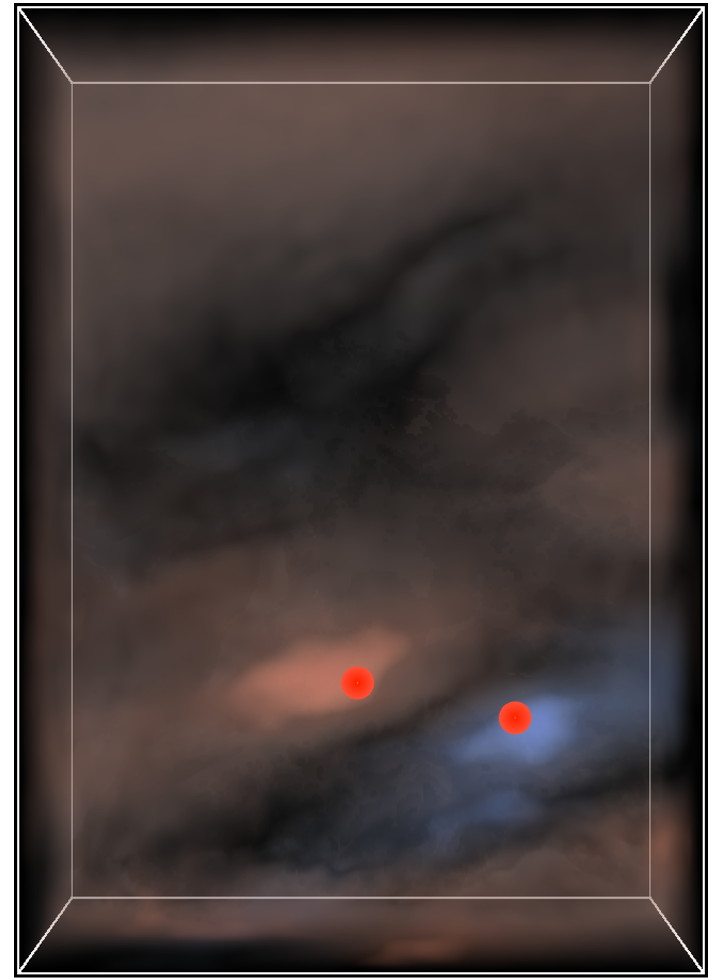}
        \includegraphics[width=0.1\textwidth]{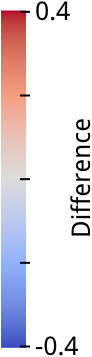}
      	\caption{$\mu = \text{\emph{u}}$, $\text{PSNR}=82.77$}
      	\label{fig:Pearson_u_multi_point_a}
    \end{subfigure}%
    \begin{subfigure}[b]{0.48\columnwidth}
      	\centering
      	\includegraphics[width=0.4\textwidth]{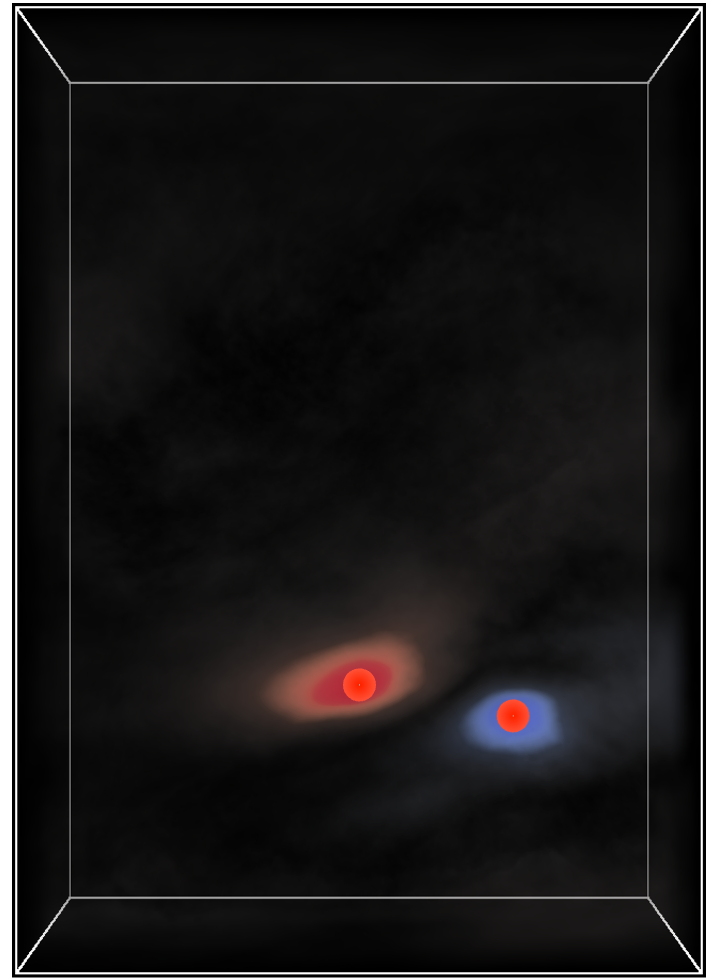}
        \includegraphics[width=0.4\textwidth]{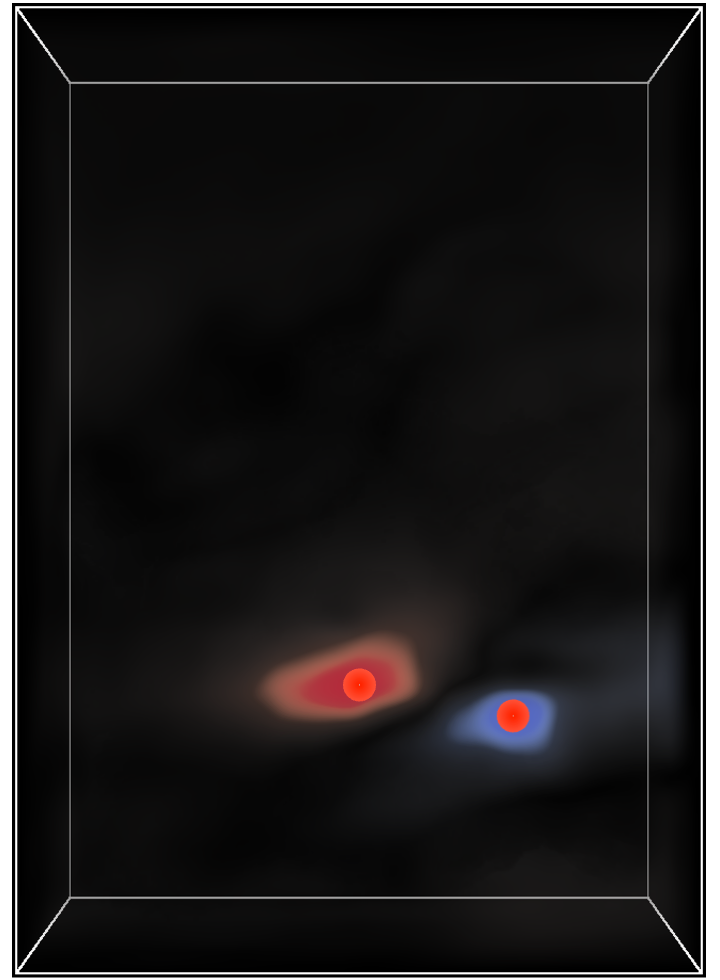}
        \includegraphics[width=0.1\textwidth]{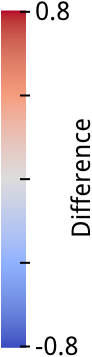}
      	\caption{$\mu = \text{\emph{u}}$, $\text{PSNR}=83.91$}
      	\label{fig:Pearson_u_multi_point_b}
    \end{subfigure}%
    \caption{Difference field visualization for multiple points in longitudinal component \emph{u} of wind speed. a) Pearson correlation field for two different points, ground truth (left), model reconstruction (right). b) MI for two different points, ground truth (left), and model reconstruction (right).}
    \label{fig:Pearson_u_multi_point}
\end{figure}

% Secondly, the user selects two points in the domain, and the difference between the reconstructed dependence fields for each point is visualized. Then, the difference fields are automatically computed and shown in Figure \ref{fig:Pearson_u_multi_point_a} for Pearson correlation and Figure \ref{fig:Pearson_u_multi_point_b} for MI. The images reveal the correlation decay around the points, as well as additional structures present in other areas within the fields. This enables users to efficiently analyze the differences in the dependence structures related to different points in the 3D domain. 
\noindent \textbf{Multi-point experiment} Secondly, the user selects two points in the domain, and the difference between the reconstructed dependence fields for each point is visualized (Figure \ref{fig:Pearson_u_multi_point_a} for Pearson correlation and Figure \ref{fig:Pearson_u_multi_point_b} for MI). The images show the correlation decay around the points and reveal additional structures in other areas within the fields. This allows users to efficiently analyze the differences in the dependence structures related to different points in the 3D domain.

\noindent \textbf{Multi-variable experiment} Our last experiment demonstrates the use of NDFs for analyzing spatial dependencies between different variables. For a set of $d\in \mathbb{N}$ variable fields, i.e.,\ $V = \{\nu_1, \nu_2, ..., \nu_d\}$, this requires training of $d (d + 1) / 2$ NDFs $\Phi_{\nu_i\nu_j}$, for each of the combinations $\nu_i, \nu_j \in V$ with $1 \le i \le j \le d$. For a pair of two different physical variables, such as $\nu_1 = \text{\emph{tk}}$ and $\nu_2 = \text{\emph{u}}$, this amounts to training three NDFs, $\Phi_{\text{\emph{tk}},\text{\emph{tk}}}$, $\Phi_{\text{\emph{u}},\text{\emph{u}}}$ and $\Phi_{\text{\emph{tk}},\text{\emph{u}}}$, which emulate the corresponding correlation fields. Note that no separate model is required for $\Phi_{\text{\emph{u}},\text{\emph{tk}}}$ if the underlying correlation measure $\rho$ (see section \ref{sec:measures}) is symmetric under exchange of arguments, since due to the symmetric architecture of NDFs $\Phi_{\text{\emph{u}},\text{\emph{tk}}}(\mathbf{p}_1, \mathbf{p}_2) = \Phi_{\text{\emph{tk}},\text{\emph{u}}}(\mathbf{p}_2, \mathbf{p}_1)$ for all $\mathbf{p}_1,\mathbf{p}_2\in\Omega$.

\begin{figure}%[t]
    \centering
    \begin{subfigure}[b]{0.48\columnwidth}
      	\centering
      	\includegraphics[width=\textwidth]{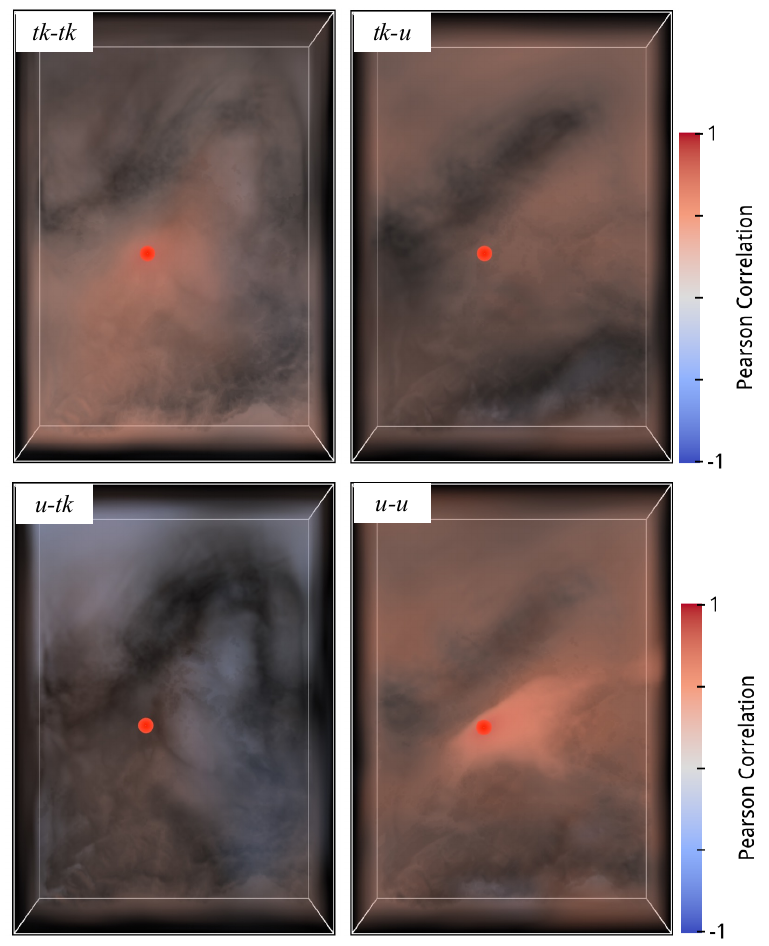}
      	\caption{Ground truth}
    %   	\label{}
    \end{subfigure}%
    \begin{subfigure}[b]{0.48\columnwidth}
      	\centering
      	\includegraphics[width=\textwidth]{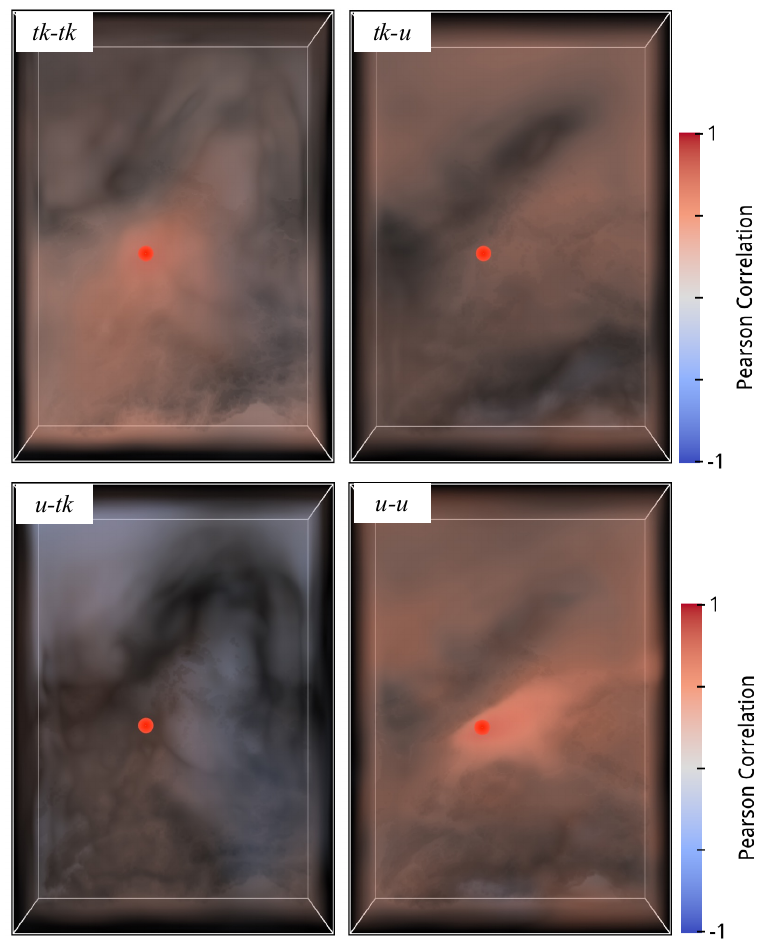}
      	\caption{NDF reconstruction}
    %   	\label{}
    \end{subfigure}%
    \caption{
    Visualizing dependencies between variables temperature (\emph{tk}) and longitudinal component \emph{u} of wind speed using Pearson correlation coefficients. a) Ground truth, b) NDF model reconstruction. Fields show correlations between variables \emph{tk} or \emph{u} at selected points to the same or different variable at other points.
    % \emph{tk}-\emph{tk} field shows correlations between selected point in \emph{tk} to all other points in the same variable. \emph{tk}-\emph{u} field shows correlations between selected point in \emph{tk} to all points in \emph{u}. \emph{u}-\emph{tk} field shows correlations between selected point in \emph{u} to all points in \emph{tk}. \emph{u}-\emph{u} field shows correlations between selected point in \emph{u} to all other points in the same variable.
    }
    \label{fig:Pearson tk u}
\end{figure}

Elaborating on the example of \emph{tk} and \emph{u}, we propose a matrix-like arrangement of linked volumetric correlation visualizations in the spirit of standard correlation matrix visualizations, i.e., \ correlation volume matrices. For this, a single reference point is selected, and correlation fields with respect to this point are rendered for all NDF configurations, i.e., \ all combinations of variables.  
Figure \ref{fig:Pearson tk u} shows an example of this. 
NDFs enable easier visualization with multiple variables as they reduce computation time and memory usage compared to on-the-fly computations using raw data. The standard method would require all variables' data in GPU memory, and even high-end GPUs with 24\,GB of memory can only load two variable fields simultaneously in practice.
%On our setup on a PC and two high-resolution monitors, about 1.4 GB of graphics memory is reserved by the system for the desktop environment display. Additionally, multiple render targets and local memory for computations further affect the available memory budget.
% With NDFs of 1\,GB each, networks for up to four variables can easily fit into the same memory. For even larger numbers of variables, it would be interesting to investigate the reuse of the variable-specific encoder grids in different networks to prevent the memory requirements of NDFs from growing quadratically with the number of variables.
NDFs of 1\,GB each allow networks for up to four variables to fit into the same memory. Exploring the reuse of variable-specific encoder grids in different networks could prevent the memory requirements of NDFs from growing quadratically with the number of variables.

%-------------------------------------------------------------------------

\section{Conclusion and Future Work\label{sec:conclusion}}

% We have introduced and evaluated neural dependence fields (NDFs) -- a novel approach for encoding and visualizing statistical dependencies in large 3D ensemble fields. NDFs are an effective means for inferring spatial dependencies in a single and between pairs of different variables. NDFs provide compact representations of linear and non-linear dependence patterns in large ensembles, enabling fast reconstruction of correlation samples directly from the compact representation. By embedding NDFs into GPU-accelerated direct volume ray-casting, we have demonstrated interactive visual analysis of 3D dependence structures.
We have introduced and evaluated neural dependence fields (NDFs), a novel approach for encoding and visualizing statistical dependencies in large 3D ensemble fields. NDFs infer spatial dependencies within single variables and in pairs of different variables. They offer compact representations of linear and non-linear dependence patterns in large ensembles, facilitating the rapid reconstruction of correlation samples from the compact representation. We demonstrated interactive visual analysis of 3D dependence structures through GPU-accelerated direct volume rendering.
%This cannot be achieved with alternative approaches. With a large weather forecast ensemble, we have shown that NDFs can reduce memory requirements and reconstruction times tremendously, especially for non-linear dependence measures, such as mutual information. Going beyond single-parameter self-correlations, the gain in execution speed opens up the opportunity to analyze multi-variable correlation patterns within their spatial context through linked volumetric views arranged in the form of correlation volume matrices.

% Our evaluation has shown that NDFs can faithfully encode and reconstruct the most prominent dependence structures in 3D fields while smoothing out details due to limited network capacities. Thus, in the future, we will investigate means to extend these capacities by adding additional network stages in both the encoder and decoder, exploring alternative network architectures like diffusion networks, and considering alternative loss functions to preserve fine details better, e.g., gradient regularization \cite{lu2021cnr}. 
% Even though we have not considered NDFs for inferring temporal dependence structures in time-varying ensemble fields, a straightforward generalization of NDFs towards this use case consists in decomposing time-varying variable data into independent fields with only spatial variation. 
% We plan to add support for time-varying data in the future, which, at the same time, would further allow for additional application scenarios in scientific workflows, such as ensemble sensitivity analysis~\cite{kumpf2018visual}.
Our evaluations show that NDFs faithfully encode and reconstruct the prominent dependence structures in 3D fields while smoothing out some details due to limited network capacities. In the future, we aim to enhance these capacities by adding network stages in the encoder and decoder, exploring alternative architectures like diffusion networks, and trying different loss functions for preserving fine details better (e.g., gradient regularization \cite{lu2021cnr}). Additionally, we plan to extend NDFs to infer temporal dependence structures in time-varying ensemble fields by decomposing the data into independent fields with spatial variation. This extension would enable new application scenarios in scientific workflows, such as ensemble sensitivity analysis \cite{kumpf2018visual}.

% In addition to plain correlation encoding, the proposed architecture, trained with the objective in Figure \ref{eq:optim}, can be seen as a tool for learning representations of the ensemble variables in an unsupervised manner. For this, a group of NDFs would be trained jointly on all variables, and a suitable target correlation measure would provide a fast-to-compute unsupervised learning signal. Considering the merger stage as an information bottleneck, see, e.g. \cite{tishby2000information,alemi2016deep}, the output fields of the variable-specific encoders can then be seen as an ensemble representation of reduced dimensionality, which preserves spatial and inter-variable correlations as closely as possible. Such representations can be interesting for generating ensemble summaries, member clustering, and uncertainty visualization and will be investigated in future work.   

\subsubsection*{Acknowledgments}
The research leading to these results has been done within the subproject “A7” of the Transregional Collaborative Research Center SFB / TRR 165 “Waves to Weather” (\url{www.wavestoweather.de}) funded by the German Research Foundation (DFG). The
authors acknowledge crucial contributions by Juan Ruiz and the RIKEN Data Assimilation Research
Team for conducting the 1000-member ensemble simulation.

%-------------------------------------------------------------------------
% \subsection{References}

% List all bibliographical references in 9-point Times, single-spaced, at the
% end of your paper in alphabetical order. 
% When referenced in the text, enclose
% the citation index in square brackets, 
% \cite{Lous90}. 
% Where
% appropriate, include the name(s) of editors of referenced books.

% For your references please use the following algorithm:

%-------------------------------------------------------------------------

% \bibliographystyle{eg-alpha}
\bibliographystyle{eg-alpha-doi}

\bibliography{main-VMV2023-sub}

\newcommand{\etalchar}[1]{$^{#1}$}
\begin{thebibliography}{\uppercase{FKMW{\etalchar{*}}23}}

\bibitem[BBR{\etalchar{*}}18]{MINE}
\textsc{Belghazi M.~I., Baratin A., Rajeshwar S., Ozair S., Bengio Y.,
  Courville A., Hjelm D.}:
\newblock Mutual information neural estimation.
\newblock In \emph{Proceedings of the 35th International Conference on Machine
  Learning} (10--15 Jul 2018), Dy J., Krause A., (Eds.), vol.~80 of
  \emph{Proceedings of Machine Learning Research}, PMLR, pp.~531--540.

\bibitem[BDSW13]{Biswas2013}
\textsc{Biswas A., Dutta S., Shen H.-W., Woodring J.}:
\newblock An information-aware framework for exploring multivariate data sets.
\newblock \emph{IEEE Transactions on Visualization and Computer Graphics 19},
  12 (2013), 2683--2692.
\newblock \href {https://doi.org/10.1109/TVCG.2013.133}
  {\path{doi:10.1109/TVCG.2013.133}}.

\bibitem[BMLC19]{Berenjkoub2019}
\textsc{Berenjkoub M., Monico R.~O., Laramee R.~S., Chen G.}:
\newblock Visual analysis of spatia-temporal relations of pairwise attributes
  in unsteady flow.
\newblock \emph{IEEE Transactions on Visualization and Computer Graphics 25}, 1
  (2019), 1246--1256.
\newblock \href {https://doi.org/10.1109/TVCG.2018.2864817}
  {\path{doi:10.1109/TVCG.2018.2864817}}.

\bibitem[CLI{\etalchar{*}}20]{localsdf2020}
\textsc{Chabra R., Lenssen J.~E., Ilg E., Schmidt T., Straub J., Lovegrove S.,
  Newcombe R.}:
\newblock Deep local shapes: Learning local sdf priors for detailed 3d
  reconstruction.
\newblock In \emph{Computer Vision--ECCV 2020: 16th European Conference,
  Glasgow, UK, August 23--28, 2020, Proceedings, Part XXIX 16} (2020),
  Springer, pp.~608--625.

\bibitem[CT{\etalchar{*}}91]{cover1991entropy}
\textsc{Cover T.~M., Thomas J.~A., et~al.}:
\newblock Entropy, relative entropy and mutual information.
\newblock \emph{Elements of information theory 2}, 1 (1991), 12--13.

\bibitem[CWMW11]{Chen2011}
\textsc{Chen C.-K., Wang C., Ma K.-L., Wittenberg A.~T.}:
\newblock Static correlation visualization for large time-varying volume data.
\newblock In \emph{2011 IEEE Pacific Visualization Symposium} (2011),
  pp.~27--34.
\newblock \href {https://doi.org/10.1109/PACIFICVIS.2011.5742369}
  {\path{doi:10.1109/PACIFICVIS.2011.5742369}}.

\bibitem[CXG{\etalchar{*}}22]{chen2022tensorf}
\textsc{Chen A., Xu Z., Geiger A., Yu J., Su H.}:
\newblock Tensorf: Tensorial radiance fields.
\newblock In \emph{Computer Vision--ECCV 2022: 17th European Conference, 2022,
  Proceedings, Part XXXII} (2022), Springer, pp.~333--350.

\bibitem[CZ19]{srn2019chen}
\textsc{Chen Z., Zhang H.}:
\newblock Learning implicit fields for generative shape modeling.
\newblock In \emph{Proceedings of the IEEE/CVF Conference on Computer Vision
  and Pattern Recognition} (2019), pp.~5939--5948.

\bibitem[EHL21]{Evers2021}
\textsc{Evers M., Huesmann K., Linsen L.}:
\newblock {Uncertainty-aware Visualization of Regional Time Series Correlation
  in Spatio-temporal Ensembles}.
\newblock \emph{Computer Graphics Forum} (2021).
\newblock \href {https://doi.org/10.1111/cgf.14326}
  {\path{doi:10.1111/cgf.14326}}.

\bibitem[FH23]{NDFZenodo}
\textsc{Farokhmanesh F., Höhlein K.}:
\newblock {Neural Fields for Interactive Visualization of Statistical
  Dependencies in 3D Simulation Ensembles: Code for Experiments}, July 2023.
\newblock \href {https://doi.org/10.5281/zenodo.8186686}
  {\path{doi:10.5281/zenodo.8186686}}.

\bibitem[FKMW{\etalchar{*}}23]{fridovich2023k}
\textsc{Fridovich-Keil S., Meanti G., Warburg F., Recht B., Kanazawa A.}:
\newblock K-planes: Explicit radiance fields in space, time, and appearance.
\newblock \emph{arXiv preprint arXiv:2301.10241} (2023).

\bibitem[GW10]{gu2010study}
\textsc{Gu Y., Wang C.}:
\newblock A study of hierarchical correlation clustering for scientific volume
  data.
\newblock In \emph{Advances in Visual Computing: 6th International Symposium,
  ISVC 2010, 1, 2010, Proceedings, Part III 6} (2010), Springer, pp.~437--446.

\bibitem[HSB{\etalchar{*}}20]{hoang2020}
\textsc{Hoang D., Summa B., Bhatia H., Lindstrom P., Klacansky P., Usher W.,
  Bremer P.-T., Pascucci V.}:
\newblock Efficient and flexible hierarchical data layouts for a unified
  encoding of scalar field precision and resolution.
\newblock \emph{IEEE Transactions on Visualization and Computer Graphics 27}, 2
  (2020), 603--613.

\bibitem[HWW22]{hoehlein2022}
\textsc{H{\"o}hlein K., Weiss S., Westermann R.}:
\newblock Evaluation of volume representation networks for meteorological
  ensemble compression.

\bibitem[KB14]{adam}
\textsc{Kingma D.~P., Ba J.}:
\newblock Adam: A method for stochastic optimization.
\newblock \emph{arXiv preprint arXiv:1412.6980} (2014).

\bibitem[KRRW19]{kumpf2018visual}
\textsc{Kumpf A., Rautenhaus M., Riemer M., Westermann R.}:
\newblock Visual analysis of the temporal evolution of ensemble forecast
  sensitivities.
\newblock \emph{IEEE Transactions on Visualization and Computer Graphics 25}, 1
  (2019).
\newblock \href {https://doi.org/10.1109/TVCG.2018.2864901}
  {\path{doi:10.1109/TVCG.2018.2864901}}.

\bibitem[KSG04]{KraskovMI}
\textsc{Kraskov A., St\"ogbauer H., Grassberger P.}:
\newblock Estimating mutual information.
\newblock \emph{Phys. Rev. E 69} (Jun 2004), 066138.
\newblock \href {https://doi.org/10.1103/PhysRevE.69.066138}
  {\path{doi:10.1103/PhysRevE.69.066138}}.

\bibitem[LJLB21]{lu2021cnr}
\textsc{Lu Y., Jiang K., Levine J.~A., Berger M.}:
\newblock Compressive neural representations of volumetric scalar fields.
\newblock In \emph{Computer Graphics Forum} (2021), vol.~40, Wiley Online
  Library, pp.~135--146.

\bibitem[LS16]{Liu2016}
\textsc{Liu X., Shen H.-W.}:
\newblock Association analysis for visual exploration of multivariate
  scientific data sets.
\newblock \emph{IEEE Transactions on Visualization and Computer Graphics 22}, 1
  (2016), 955--964.
\newblock \href {https://doi.org/10.1109/TVCG.2015.2467431}
  {\path{doi:10.1109/TVCG.2015.2467431}}.

\bibitem[LWS18]{Liebmann2018}
\textsc{Liebmann T., Weber G.~H., Scheuermann G.}:
\newblock Hierarchical correlation clustering in multiple 2d scalar fields.
\newblock \emph{Computer Graphics Forum 37}, 3 (2018), 1--12.
\newblock \href {https://doi.org/https://doi.org/10.1111/cgf.13396}
  {\path{doi:https://doi.org/10.1111/cgf.13396}}.

\bibitem[MESK22]{MultiresHashEncoding}
\textsc{M\"{u}ller T., Evans A., Schied C., Keller A.}:
\newblock Instant neural graphics primitives with a multiresolution hash
  encoding.
\newblock \emph{ACM Trans. Graph. 41}, 4 (jul 2022).
\newblock \href {https://doi.org/10.1145/3528223.3530127}
  {\path{doi:10.1145/3528223.3530127}}.

\bibitem[MLL{\etalchar{*}}21]{martel2021acorn}
\textsc{Martel J.~N., Lindell D.~B., Lin C.~Z., Chan E.~R., Monteiro M.,
  Wetzstein G.}:
\newblock Acorn: Adaptive coordinate networks for neural scene representation.
\newblock \emph{arXiv preprint arXiv:2105.02788} (2021).

\bibitem[MON{\etalchar{*}}19]{occupancy2019}
\textsc{Mescheder L., Oechsle M., Niemeyer M., Nowozin S., Geiger A.}:
\newblock Occupancy networks: Learning 3d reconstruction in function space.
\newblock In \emph{Proceedings of the IEEE/CVF conference on computer vision
  and pattern recognition} (2019), pp.~4460--4470.

\bibitem[MRL95]{KDEMI}
\textsc{Moon Y.-I., Rajagopalan B., Lall U.}:
\newblock Estimation of mutual information using kernel density estimators.
\newblock \emph{Physical Review E 52}, 3 (1995), 2318.

\bibitem[MRNK21a]{muller2021}
\textsc{M{\"u}ller T., Rousselle F., Nov{\'a}k J., Keller A.}:
\newblock Real-time neural radiance caching for path tracing.
\newblock \emph{arXiv preprint arXiv:2106.12372} (2021).

\bibitem[MRNK21b]{FusedMLP}
\textsc{M\"{u}ller T., Rousselle F., Nov\'{a}k J., Keller A.}:
\newblock Real-time neural radiance caching for path tracing.
\newblock \emph{ACM Trans. Graph. 40}, 4 (jul 2021).
\newblock \href {https://doi.org/10.1145/3450626.3459812}
  {\path{doi:10.1145/3450626.3459812}}.

\bibitem[MST{\etalchar{*}}21]{mildenhall2021nerf}
\textsc{Mildenhall B., Srinivasan P.~P., Tancik M., Barron J.~T., Ramamoorthi
  R., Ng R.}:
\newblock Nerf: Representing scenes as neural radiance fields for view
  synthesis.
\newblock \emph{Communications of the ACM 65}, 1 (2021), 99--106.

\bibitem[Mü21]{tiny-cuda-nn}
\textsc{Müller T.}:
\newblock {tiny-cuda-nn}, 4 2021.
\newblock URL: \url{https://github.com/NVlabs/tiny-cuda-nn}.

\bibitem[NGW{\etalchar{*}}20]{Necker2020}
\textsc{Necker T., Geiss S., Weissmann M., Ruiz J., Miyoshi T., Lien G.-Y.}:
\newblock A convective-scale 1,000-member ensemble simulation and potential
  applications.
\newblock \emph{Quarterly Journal of the Royal Meteorological Society 146}, 728
  (2020), 1423--1442.
\newblock \href {https://doi.org/https://doi.org/10.1002/qj.3744}
  {\path{doi:https://doi.org/10.1002/qj.3744}}.

\bibitem[NS23]{CorrerenderZenodo}
\textsc{Neuhauser C., Stumpfegger J.}:
\newblock {chrismile/Correrender}: A correlation field renderer using the
  {Vulkan} graphics {API}, v2023-07-29, July 2023.
\newblock \href {https://doi.org/10.5281/zenodo.8195623}
  {\path{doi:10.5281/zenodo.8195623}}.

\bibitem[PFS{\etalchar{*}}19]{park2019deepsdf}
\textsc{Park J.~J., Florence P., Straub J., Newcombe R., Lovegrove S.}:
\newblock Deepsdf: Learning continuous signed distance functions for shape
  representation.
\newblock In \emph{Proceedings of the IEEE/CVF conference on computer vision
  and pattern recognition} (2019), pp.~165--174.

\bibitem[PW12]{Pfaffelmoser2012}
\textsc{Pfaffelmoser T., Westermann R.}:
\newblock Visualization of global correlation structures in uncertain 2d scalar
  fields.
\newblock In \emph{Computer Graphics Forum} (2012), vol.~31, Wiley Online
  Library, pp.~1025--1034.

\bibitem[STS06]{Sauber2006}
\textsc{Sauber N., Theisel H., Seidel H.-P.}:
\newblock Multifield-graphs: An approach to visualizing correlations in
  multifield scalar data.
\newblock \emph{IEEE Transactions on Visualization and Computer Graphics 12}, 5
  (2006), 917--924.
\newblock \href {https://doi.org/10.1109/TVCG.2006.165}
  {\path{doi:10.1109/TVCG.2006.165}}.

\bibitem[TFT{\etalchar{*}}20]{tewari2020nr}
\textsc{Tewari A., Fried O., Thies J., Sitzmann V., Lombardi S., Sunkavalli K.,
  Martin-Brualla R., Simon T., Saragih J., Nie{\ss}ner M., et~al.}:
\newblock State of the art on neural rendering.
\newblock In \emph{Computer Graphics Forum} (2020), vol.~39, Wiley Online
  Library, pp.~701--727.

\bibitem[{The}23]{VulkanSpec}
\textsc{{The Khronos Vulkan Working Group}}:
\newblock {Vulkan 1.3.245 - A Specification}.
\newblock
  \url{https://registry.khronos.org/vulkan/specs/1.3-extensions/html/vkspec.html},
  2023.
\newblock Accessed: 2023-03-30.

\bibitem[TLY{\etalchar{*}}21]{takikawa2021}
\textsc{Takikawa T., Litalien J., Yin K., Kreis K., Loop C., Nowrouzezahrai D.,
  Jacobson A., McGuire M., Fidler S.}:
\newblock Neural geometric level of detail: Real-time rendering with implicit
  3d shapes.
\newblock In \emph{Proceedings of the IEEE/CVF Conference on Computer Vision
  and Pattern Recognition} (2021), pp.~11358--11367.

\bibitem[TSM{\etalchar{*}}20]{tancik2020}
\textsc{Tancik M., Srinivasan P., Mildenhall B., Fridovich-Keil S., Raghavan
  N., Singhal U., Ramamoorthi R., Barron J., Ng R.}:
\newblock Fourier features let networks learn high frequency functions in low
  dimensional domains.
\newblock \emph{Advances in Neural Information Processing Systems 33} (2020),
  7537--7547.

\bibitem[WHW22]{weiss2022fvsrn}
\textsc{Weiss S., Herm{\"u}ller P., Westermann R.}:
\newblock Fast neural representations for direct volume rendering.
\newblock In \emph{Computer Graphics Forum} (2022), vol.~41, Wiley Online
  Library, pp.~196--211.

\bibitem[XTS{\etalchar{*}}22]{xieneural}
\textsc{Xie Y., Takikawa T., Saito S., Litany O., Yan S., Khan N., Tombari F.,
  Tompkin J., Sitzmann V., Sridhar S.}:
\newblock Neural fields in visual computing and beyond.
\newblock In \emph{Computer Graphics Forum} (2022), vol.~41, Wiley Online
  Library, pp.~641--676.

\bibitem[ZD11]{CopulaMI}
\textsc{Zeng X., Durrani T.}:
\newblock Estimation of mutual information using copula density function.
\newblock \emph{Electronics letters 47}, 8 (2011), 493--494.

\bibitem[ZMZM14]{zhangvisual}
\textsc{Zhang Z., McDonnell K.~T., Zadok E., Mueller K.}:
\newblock Visual correlation analysis of numerical and categorical data on the
  correlation map.
\newblock \emph{IEEE transactions on visualization and computer graphics 21}, 2
  (2014), 289--303.

\end{thebibliography}

%-------------------------------------------------------------------------

\end{document}